\documentclass[11pt,letterpaper,twoside]{article}
\usepackage[nohead,marginratio={1:1,1:1},vscale=0.8]{geometry}
\usepackage{graphicx}
\usepackage{caption} 
\usepackage{float} 

\author{Frederik Eaton} 
\date{First preprint: 22 November, 2024 \\
Latest revision: 18 March, 2025}
\title{Influence functions and regularity tangents \\
  for efficient active learning}

\usepackage{amsfonts}
\usepackage{amsmath}
\newcommand{\T}{\top}
\newcommand{\bb}[1]{\mathbb{ #1 }}
\newcommand{\R}{\bb{R}}
\newcommand{\norm}[1]{\lVert #1 \rVert}
\newcommand{\abs}[1]{\left| #1 \right|}
\renewcommand{\(}{\left(}
\renewcommand{\)}{\right)}
\newcommand{\argmax}{\mathop{\mathrm{argmax}}}
\newcommand{\argmin}{\mathop{\mathrm{argmin}}}
\renewcommand{\L}{\mathcal{L}}
\newcommand{\partby}[2]{{\partial #1\over \partial #2}}
\newcommand{\partbyby}[3]{{\partial^2 #1\over \partial #2 \partial #3}}
\newcommand{\partbyt}[2]{{\partial^2 #1\over {\partial #2}^2}}
\newcommand{\ud}{\mathrm{d}}
\newcommand{\dby}[2]{\frac{\ud #1}{\ud #2}}
\newcommand{\eps}{\varepsilon}
\newcommand{\ditto}{\texttt{"}}
\newcommand{\I}{\mathcal{I}}
\newcommand{\Iup}{\I_{\textrm{up,params}}}
\newcommand{\Iul}{\I_{\textrm{up,loss}}}
\newcommand{\Iur}{\I_{\textrm{up,reg}}}
\newcommand{\zt}{z_{\textrm{test}}}
\newcommand{\zu}{z}
\newcommand{\gl}{\sigma}
\newcommand{\E}{\bb{E}}
\renewcommand{\l}{l} 
\newcommand{\h}{h} 
\renewcommand{\Gamma}{\textrm{Gamma}}
\newcommand{\where}{\:\big|\:}

\newcommand{\dottheta}{\dot{\theta}}

\newcommand{\eqntitle}[1]{
\vspace{0.8em}\newline
\indent\parbox{\textwidth}{
  \indent\textsc{(#1)}
}\nopagebreak\vspace{-0.6em}}

\begin{document}
\maketitle

\section*{Abstract}

In this paper we describe an efficient method for providing a
regression model with a sense of curiosity about its data. In the
field of machine learning, our framework for representing curiosity is
called {\em active learning}, which concerns the problem of
automatically choosing data points for which to query labels in the
semi-supervised setting. The methods we propose are based on computing
a ``regularity tangent'' vector that can be calculated (with only a
constant slow-down) together with the model's parameter vector during
training. We then take the inner product of this tangent vector with
the gradient vector of the model's loss at a given data point to
obtain a measure of the influence of that point on the complexity of
the model. In the simplest instantiation, there is only a single
regularity tangent vector, of the same dimension as the parameter
vector. Thus, in the proposed technique, once training is complete,
evaluating our ``curiosity'' about a potential query data point can be
done as quickly as calculating the model's loss gradient at that
point. The new vector only doubles the amount of storage required by
the model. We show that the quantity computed by our technique is an
example of an ``influence function'', and that it measures the
expected squared change in model complexity incurred by up-weighting a
given data point. We propose a number of ways for using this and other
related quantities to choose new training data points for a regression
model.

\tableofcontents

\section{Background} \label{sec:back}

The sub-discipline of computer science called machine learning is
concerned with modeling observations of any kind, whether
``real-world'' or simulated. It is used in many important applications
today, including everything from AI language models to search engines
to geographical models. The input to a machine learning problem
typically consists of a sequence of ``exchangeable'' observations,
which is to say that the ordering of the observations is considered
irrelevant. A typical approach to modeling these observations entails
searching for values for some numerical (i.e.~floating point)
variables or {\em parameters} controlling the predictions of a
probabilistic symbolic model of the input, such that the predicted
probability of the set of observations is maximized. Equivalently, a
parameter vector may be sought that minimizes some additive measure of
model error. This is called {\em regression}, and it seems fair to say
that most machine learning applications are based on one or more
applications of some regression
formalism.

The input to a regression problem often consists of a set of data points with
human-generated or human-curated labels. In this context the labels are
usually more expensive to
procure than the unlabeled data points. It has long been recognized
that machine learning models in many applications
could be trained more efficiently by
giving them a way to choose unlabeled data points for which to query
new labels, given that in general not all data points will be equally informative to
the model. We propose a new method for solving this problem, the
primary strength of which is efficiency, as the method can be used
with even the very largest models, and does not add to the time
complexity of regression. The aforementioned problem of finding data
points for which to query labels is called {\em active learning}. There have
been attempts to design efficient active learning query selectors
using {\em influence functions}, which measure the derivatives of some
computed quantity such as model parameters with respect to the
infinitesimal up-weighting of a data point. We show that our proposed
method is an example of the use of an influence function, which in our
case estimates the effect of a candidate data point on one measure of
model complexity (called the {\em regularizer}, which is often already
a part of the model). We also show that an established algorithm for
computing influence functions called LiSSA \cite{agarwal2017b,koh2017}
is a special case of an algorithm we propose for our method, called
{\em stochastic gradient descent with forward-mode automatic
  differentiation}, or SGDF.

There is a good deal of published research which touches on some of
the ideas in this paper, yet although the algorithm we propose is
very simple, we have not discovered where it has been proposed before. The rest
of this section gives a review of some relevant background in the
published Machine Learning literature, which might make more sense
after reading sections \ref{sec:defs} and \ref{sec:obs}.

We do not know whether anyone else has remarked on the fact that the
LiSSA algorithm \cite{agarwal2017b} can be seen as a special case of
what we call SGDF (section \ref{sec:sgdf}). It is recognized that one
can produce derivatives with respect to a hyperparameter by modifying
the SGD updates, and this has been used for optimization of
hyperparameters
\cite{gilbert1992,bengio1999,maclaurin2015,franceschi2017,franceschi2018,lorraine2020}
but not as far as we know for selecting data points in active
learning, aside from applications using multiple point-specific influence functions, which may require
a long computation to be executed for each data point.


Several publications have explored active learning using these
``influence functions'' \cite{cook1982}, based, as we said, upon
calculating the total derivative of the loss with respect to a
hyperparameter $\eps$ being used to ``up-weight'' a given data point $z$
\cite{koh2017,liu2021,schioppa2021,xia2023}. But to our knowledge none
of these have shown how to calculate $\dby{R_s(\theta)}{\eps}$ as the
inner product of a single regularity tangent vector
$\dby{\theta^*(s)}{s}$ and the loss gradient vector
$\partby{L}{\theta}(z,\theta)$. We are not aware of a proposal in any
of the active learning literature to use the model complexity as
expressed by the regularizer $R_s(\theta)$ for measuring model change,
and in prior publications the term ``model change'' is usually taken
to mean the change in loss on a test set, rather than measuring the
change in any simple function of the parameters such as a regularizer
term \cite{cai2013}.

Others have demonstrated how to derive linear approximations to the
parameter vectors appearing in the $k$-fold cross validation objective
function (our equations \ref{eq:gcv}-\ref{eq:loocv-opt-reg})
\cite{giordano2019}. Our formula for the approximate leave-one-out
generalization error, given in equation \ref{eq:gpert} as a ``summed
self-influence'' perturbation on the empirical risk, might be found in
\cite{golub1979} in some form.

The example objective functions that we used to
discuss active learning in a multi-user setting in section \ref{sec:reg-multi-user} were created for this
paper.

Some applications use a technique called ``query by committee'' in
which a ``committee'' of some small number of models is trained
equivalently but with different random initializations, and the
predictions of each model are combined to form a measure of the
uncertainty of the model at each point \cite{settles2010,cohn1994}.
Although such techniques can help us measure the uncertainty of our
predictions, they do not necessarily make good query selectors for
active learning. Some data points may have high uncertainty simply
because they are not possible for the model to accommodate well, and
not because they are ``interesting'' in the sense of being able to
affect the model complexity as explained below.

An early version of this paper was protected by provisional patent
application 63/551,085, which was filed with the United States Patent
and Trademark Office on February 8, 2024.

\section{Definitions and preliminaries} \label{sec:defs}

\subsection{Regression and regularization} \label{sec:reg-reg}

Consider the problem of least squares linear regression, also called
ordinary least squares (OLS), which is to minimize the function
\begin{align}
f(\theta) = {1\over n} \sum_i(\theta^\T x_i - y_i)^2 \label{eq:ols}
\end{align}
where $\theta$ is a vector of parameters in $\R^p$ and
$\{z_i=(x_i,y_i)|i=1\ldots n\}$ is a set of data points in $\R^p\times
\R$. The function $f$ is called the {\em empirical risk}. The response
variables $y_i$ may also be vector-valued, in which case the square
loss terms $(\theta^\T x_i-y_i)^2$ become $\L_2$ norms $\norm{\theta
  x_i - y_i}^2$ where $\theta$ is now a matrix. The normalization
coefficient ${1\over n}$ makes the objective an average of squared
differences, and this normalization is often desirable when
thinking about the training optimization as error minimization.
In the probabilistic (maximum likelihood) interpretation of the
optimization, normalization is omitted (see next section).
In either case it does not affect the
location of the optimum parameter vector $\theta^*$.

The linear least squares regression problem is seen as a special case of the
more general nonlinear least squares regression problem
\begin{align}
f(\theta)&={1\over n}\sum_i \norm{F(x_i;\theta)-y_i}^2 \label{eq:nonlin-reg}
\end{align}
where $F$ is a function being fit to some data points
$\{(x_i,y_i)|i\}$. Each data point pairs an input or
feature vector $x$ with an observed output or response
variable $y$, which we also call the ``label'' of $x$.
Even more generally, $f$ may be an average over
arbitrary loss functions $L$ of input-output pairings
$z=(x,y)$:
\begin{align}
f(\theta)&={1\over n}\sum_i L(z_i,\theta)
\end{align}
To avoid over-fitting, sometimes a {\em regularization term} is
introduced:
\begin{align}
f(\theta,s)&={1\over n}\(\sum_i(\theta^\T x_i - y_i)^2 + s\norm{\theta}^2\) \label{eq:l2-reg}
\end{align}
This is called {\em regularized least squares regression} (RLS).
In the general case, we would have the model training objective as
\begin{align}
f(\theta,s)&={1\over n} \(\sum_i L(z_i,\theta) + R(s,\theta)\) \label{eq:gen-reg}
\end{align}
where $R$ is a general regularizer function (which is often proportional to
$s$ as in \ref{eq:l2-reg}, in which case we may write $sR(\theta)$ for $R(s,\theta)$).

The regularizer causes the optimization to prefer simpler models
$\theta$, in the case of equation \ref{eq:l2-reg}, by penalizing those
with high $\norm{\theta}^2$ in the minimization. Thus
$R(s,\theta)$ becomes a measure of model complexity. This penalty
term also has a probabilistic interpretation, as explained in the next
section.

\subsection{The Bayesian interpretation and cross-validation} \label{sec:bayes-cv}

The Bayesian interpretation of a regression problem can be formulated
by taking the natural exponential of the negative of the empirical risk function,
which is then treated as a probability. Ideally, this results in a
{\em maximum likelihood estimation} (MLE) optimization problem, which
refers to maximizing the likelihood of some parameters $\theta$ of a
model given some data $z$:
\begin{align}
\theta^* = \argmax_\theta P(z|\theta) = \argmax_\theta \prod_j P(z_j|\theta) \label{eq:mle}
\end{align}

MLE has the nice property of being invariant to changes in the scale
or representation of its parameters $\theta$. However, sometimes we
need to place a prior on $\theta$, which might correspond for example
to the use of a regularizer term in the regression problem:
\begin{align}
\theta^* = \argmax_\theta P(z|\theta)P(\theta|s)\ud \theta \label{eq:map}
\end{align}
This is an example of a {\em maximum a posteriori} (MAP) optimization
problem because the parameter vector $\theta$ now has a probability distribution
attached to it. If we change the representation of $\theta$ but keep
the regularizer term $R(s,\theta)$ the same, then we may get a
different maximum. Nevertheless, some reparametrizations, such as
scaling $\theta$ in the linear least squares problem with $\L_2$ regularizer
(and reciprocally scaling the features $X$),
can be compensated by a
corresponding reduction in the scale of $s$, which is itself a
hyperparameter being optimized in regularized regression.
Because of this invariance property, the
use of MAP may be more defensible to Bayesians when it provides a
probabilistic interpretation to regularization problems in the
context of model selection.
(See also equation \ref{eq:reg-reparam}, which considers transforming $R$ through a monotonic function.)

Returning to simple regression
\begin{align}
f(z,\theta)=\sum_{i=1}^n L(z_i,\theta)
\end{align}
where we have omitted the normalizer ${1\over n}$ for convenience of
exposition, and to ordinary least squares regression
\begin{align}
f(z,\theta) = \sum_{i=1}^n (\theta^\T x_i - y_i)^2
\end{align}
we form a probability distribution
\begin{align}
P(z|\theta)={1 \over Z}\exp(-f(z,\theta))={1\over Z}\exp\(-\sum_{i=1}^n\(\sum_{j=1}^p \theta_j x_{ij}-y_i\)^2\)
\end{align}
where $Z$ is a normalizing constant (in more general regression
problems $Z$ may depend on $\theta$, but does not here). This is clearly a normal distribution over the response variables $y_i$, with variance ${1\over 2}$.
So, multiplying the original regression objective by some positive constant $\alpha$,
\begin{align}
f(z,\theta,\alpha) = \alpha \sum_{i=1}^n (\theta^\T x_i - y_i)^2 \label{eq:alpha-lsr}
\end{align}
yields a normal distribution with variance ${1\over 2\alpha}$. Or it
may be more convenient to say that the {\em precision} (or inverse
variance) is $2\alpha$. In this case the normalizing constant $Z$ depends
on $\alpha$, and we should add a $\log Z(\alpha)$ term to the
objective if we are going to be minimizing with respect to $\alpha$.
We can check that, up to a constant, this is $-{n\over 2}\log \alpha$:
$Z=\sigma\sqrt{2\pi}={\sqrt{2\pi}\over\sqrt{2\alpha}}=\sqrt{\pi\over\alpha}$
for each of $n$ data points, so for the whole dataset $\log Z=-{n\over
  2} \log {\alpha \over \pi} = -{n\over 2}\log \alpha + C$.

Minimizing the new objective
\begin{align}
f(z,\theta,\alpha) = \alpha \sum_{i=1}^n \ldots - {n\over 2}\log \alpha
\end{align}
with respect to $\alpha$ by setting $\partby{f}{\alpha}$ to zero gives
\begin{align}
{1\over 2\alpha} = {1\over n} \sum_{i=1}^n \(\theta^\T x_i - y_i\)^2
\end{align}
or in other words the MLE-optimum $\alpha$ corresponds to a
distribution over the response variables whose variance ${1\over
  2\alpha}$ is equal to the average error of the model, which is an
unbiased estimator for the variance of the response of the
model\footnote{An unbiased estimator for the standard deviation is the
  square root of ${n \over n-1}$ times the unbiased variance
  estimator.}. As long as it is positive, the choice of $\alpha$ does
not affect the location of the optimum $\theta$, and so we can
simplify the least squares regression problem by omitting it. It may
nevertheless be good to remember that we have done so, in case we have
to add it back, as we do briefly in section \ref{sec:reg-multi-user}
when considering regression in the multi-user setting.

The objective for regularized least squares (RLS) regression is
\begin{align}
f(z,\theta,s) = \sum_{i=1}^n \(\theta^\T x_i - y_i\)^2 + s \norm{\theta}^2 \label{eq:rls-bayes}
\end{align}
Scaling by $\alpha$, this corresponds to
\begin{align}
f(z,\theta,s,\alpha) = \alpha \sum_{i=1}^n (\theta^\T x_i - y_i)^2 + \alpha s \norm{\theta}^2 \label{eq:aas}
\end{align}
which represents the probabilistic program
\begin{align}
\theta &\gets N(0,{1\over 2\alpha s}) \label{eq:draw-theta} \\
y_i &\gets N(\theta^\T x_i,{1\over 2\alpha}) \label{eq:draw-yi}
\end{align}
The two $\alpha$ coefficients in \ref{eq:aas} can be factored out and
thus have no effect on the location of the optimal $\theta$, and so
expositions often omit the scaling factor $\alpha$ when talking about
the structure of the RLS objective function. When optimizing the
objective with respect to $\alpha$ or $s$, we should include the
normalizer terms which are functions of these variables. They had been
omitted earlier because they are not functions of $\theta$:
\begin{align}
f(z,\theta,s,\alpha) = \alpha \sum_{i=1}^n (\theta^\T x_i - y_i)^2 + \alpha s \norm{\theta}^2 -{n\over 2}\log \alpha - {p \over 2} \log \alpha s
\end{align}
This corresponds to the factorization
\begin{align}
p(z,\theta|s,\alpha) = p(z|\theta,\alpha) \cdot p(\theta|s,\alpha) \label{eq:alpha-prob-fact}
\end{align}

Maximizing this over $\theta$ is a MAP problem, as mentioned before
(equation \ref{eq:map}). By hypothesizing a continuous,
one-dimensional family of distributions over the parameter vector
$\theta \in \R^p$, regularization allows us to reduce the number of
model variables which must be treated in a non-probabilistic fashion from $p
= \dim \R^p$ down to one, namely $s$. The regularizer term can be seen
as assigning a scalar-valued index of model complexity $R(\theta)$ to
every setting of the parameters $\theta$, and is used to penalize
complex models by allocating them a lower prior probability. The
``regularity'' $s$ is proportional to $2\alpha s$, which is the inverse
variance, or {\em precision}, of the prior distribution over the
parameters in (\ref{eq:draw-theta}).

The variable $s$, which actually controls the ratio of the variances
of the two distributions, can be optimized by minimizing the average
loss $L(z,\theta^*(s))$ on a test set (or some more data-efficient variation
of this, such
as $k$-fold cross-validation). In other words we adjust $s$ to adjust the variance
of the model parameters ${1\over 2\alpha s}$ (relative to the variance
of the data ${1\over 2 \alpha}$); we do this by trying to maximize the probability
of some new data that the trained model $\theta^*$ is ignorant about.
The inner (MAP) minimization for $\theta$ includes the regularizer term and is equivalent to RLS (equation \ref{eq:rls-bayes}), but the outer (MLE) minimization for $s$ is measuring the generalization error and should not include a regularizer.
The fact that the parameter optimization is the same as RLS shows
that this problem and
its associated algorithm have a precise Bayesian interpretation given
by (\ref{eq:draw-theta}) and (\ref{eq:draw-yi}).

Including the normalizer term $-{p\over 2 \alpha}\log s$ in the outer
minimization objective for optimizing $s$ is apparently not typically
done but may change the location of the optimum $s$ and might even be
recommended as this term was not taken into account when finding
$\theta^*$. At the same time, the term's relative contribution to the
objective becomes less significant with each new test data point.
Thus, if we have enough test data then the term may be ignored. It is
not immediately obvious what effect the inclusion of this term would
have in correcting the location of the optimal regularity $s^*$, but
it seems not unreasonable to include a term in the objective for $s$
which penalizes very small regularity values as this does. The
difference may be more apparent when there are only a few data points;
but this may still be an important domain, and for example the
multi-user setting we consider in section \ref{sec:reg-multi-user}
might prioritize getting good predictions during the first few
interactions. In summary, it could potentially be argued (by a
Bayesian) that when optimizing $s$, one should use the compensated
regularizer which includes the $s$-dependent normalization constant
for $p(\theta|s)$; and it could be argued that omitting this term is
the same as postulating some non-uniform prior over $s$ in the
optimization. We wish that we had more familiarity with these
arguments.

\vspace{1em}

Turning now to the general regularization problem,
\begin{align}
f(\theta,s)=\sum_j L(z_j,\theta)+R(s,\theta) \label{eq:gen-reg-reg}
\end{align}
we explain how to choose the regularity hyperparameter $s$ using
$k$-fold cross-validation (CV). In this technique, data points are
split into $k$ groups $\{z_i \where j=1\ldots k, i\in T_j \subseteq
\{1\ldots n\}\}$. The parameter vector $\theta$ is optimized using the remaining
$k-1$ (``training'') groups when averaging the loss for data points
$T_j$ in group $j$. In most descriptions of regularized regression,
the hyperparameter $s$ is then chosen to minimize the resulting
approximate generalization error:
\begin{align}
  s^* &= \argmin_s \sum_{j=1}^k \sum_{i\in T_j} L(z_i,\theta_j^*(s)) \label{eq:k-fold-reg-opt}
\end{align}
where $\theta_j$ represents the parameters to be used for evaluating
the objective on test set $T_j$, which as we said are trained using the remaining
points:
\begin{align}
 \theta_j^*(s) &= \argmin_\theta \(\sum_{i\notin T_j} L(z_i,\theta)+R(s,\theta)\)
\end{align}

Training $\theta_j$ on a set of data points distinct from the test set
$T_j$ ensures we are computing a quantity
that measures the model's ability to generalize itself to {\em new
  data}, which is what we are usually interested in with models.
The reasoning is that by adjusting $s$ to minimize the error on
a separate set of data points from the set which
is used to optimize $\theta$, we can avoid a situation where trained
models $L(\cdot, \theta^*(s))$ adapt too closely to their training
data and thereby generalize poorly to new data. This problem is called
``overfitting'' and it typically happens when the hyperparameter $s$
is too small (see the $\L_2$ regularizer, equation \ref{eq:l2-reg}).
The opposite scenario, when $s$ is too large and $\theta^*(s)$ fails
to capture certain patterns in the data, is called ``underfitting''.
We introduce the term {\em fit regularity} or just {\em regularity} to
refer to any scalar measurement of the degree of over or under fitting
of a model, with positive regularity corresponding to underfitting and
negative to overfitting. A hyperparameter, such as $s$ above, controlling fit
regularity is called a {\em regularity hyperparameter}.

As long as one can postulate a measure of model complexity
$R(\theta)$, it is possible to construct a regularization term
$R(s,\theta)=sR(\theta)$, i.e.~by multiplying this complexity measure with a
regularity hyperparameter $s$. Moreover, as long as it is a smooth
function of the parameters $\theta$, it seems that only the ordering
induced by $R$ on $\R^p \ni \theta$ is important; we can reparametrize the
complexity measure by applying some increasing function $t$:
\begin{align}
\tilde{R}(\theta) = t(R(\theta)) \label{eq:reg-reparam}
\end{align}
where $\tilde{R}$ is the reparametrized measure, and then the optimal $\tilde{\theta}^*(s)$ will correspond to the point where
\begin{align}
  0=\partby{}{\theta}f(\theta,s) = \sum_j \partby{L}{\theta}(z_j,\theta)+\( s\partby{\tilde{R}}{\theta}(\theta) = st'(R(\theta))\partby{R}{\theta} \)
  \label{eq:reparam-zero}
\end{align}
where $t'$ represents the derivative of $t$, and so we see that the
same optimal parameter vector $\theta^*$ is obtained at a new setting
of $s$:
\begin{align}
\tilde{\theta}^*(s) = \theta^*(st')
\end{align}
or in other words we should take the original setting of $s$ and
divide by $t'(R(\theta^*))$ to get the proper value for $s$ when
working with the new regularizer $\tilde{R}$. This holds because both
optima satisfy equation \ref{eq:reparam-zero}.

According to the ``Minimum Description Length'' model selection
principle, we should assign a higher probability to models which are
easier to describe in some pre-determined language. In this view, the
negative log probability $-\log \( P(\theta)P(z|\theta) \)$
corresponds to the length of some encoding of the data $z$ using a
model parametrized by some vector $\theta$ (the length of this being $-\log
P(z|\theta)$), with an encoded description of the model prepended to
the encoding (of length $-\log P(\theta)$). The regularizer term can
then be seen as selecting an encoding for the parameters $\theta$ by
specifying a probability distribution over them.

\subsection{Gradient descent and stochastic gradient descent}
\label{sec:gd}
Optimization of the empirical risk function $f$ can be done using
matrix arithmetic in ordinary least squares regression, but with more
general loss functions it is usually performed using {\em gradient
  descent}:
\eqntitle{Gradient Descent}
\begin{align}
\theta_{t+1} &\gets \theta_t - \eta_t \partby{f}{\theta}(\theta_t) \label{eq:gd}
\end{align}
where $\eta_t$ is a step size that typically varies from one iteration
to the next according to some schedule heuristic. Many algorithms make small adjustments to $\eta$ by
tracking the movement of $\theta$ during the algorithm (as in Adam or Adagrad \cite{kingma2015}).

For problems with large numbers of data points, the gradient descent
updates can be modified to only consider one point, or a small batch
of points, at a time. This is called {\em stochastic gradient descent}
(SGD):
\eqntitle{Stochastic Gradient Descent}
\begin{align}
\theta_{t+1} &\gets \theta_t-\eta_t \partby{L}{\theta}(z_{u_t},\theta_t) \label{eq:sgd}
\end{align}
where $u_t \in \{1\ldots n\}$ typically cycles through the indices of
the data points in a random or pseudo-random order. The regularization
term can be taken into account at each iteration:
\eqntitle{SGD, continuous regularization}
\begin{align}
\theta_{t+1} &\gets \theta_t-\eta_t \(\partby{L}{\theta}(z_{u_t},\theta_t) + {1\over n}\partby{R}{\theta}(s,\theta_t)\)  \label{eq:sgd-reg-iter}
\end{align}
or intermittently, such as at the end of each batch of $n$ points:
\eqntitle{SGD, intermittent regularization}
\begin{align}
\theta_{t+1} &\gets \theta_t-\eta_t \partby{R}{\theta}(s,\theta_t) \label{eq:sgd-reg-batch}
\end{align}
In either case, since this update usually has the effect of shrinking the
parameters uniformly, it is sometimes called parameter shrinkage. For example, when $R=s\norm{\theta}^2$, we have
\begin{align*}
\partby{R}{\theta}&=2s\theta \\
\theta_{t+1} &= \theta_t-\eta_t 2 s \theta_t \\
&= \theta_t (1-2s\eta_t)
\end{align*}

Often the regularization term is omitted or ignored, and overfitting
is avoided by stopping SGD early, for example by periodically
evaluating the model's performance on a test dataset and noticing when
this performance stops improving.\cite{bengio2012} Thus, machine
learning applications which do not explicitly invoke a regularization
formalism will typically still rely on some implicit form of
regularization in order to avoid overfitting.

\subsection{Influence functions}
Machine learning engineers may be interested in estimating the
sensitivity of the optimum $\theta^*$ to small changes in the
objective function $f$, for example if one data point is given
slightly more weight.\footnote{A recommended background paper
about influence functions is Koh and Liang, 2017, ``Understanding
Black-box Predictions via Influence Functions'' \cite{koh2017}.}
\begin{align}
f(\theta,z,\eps) &= \sum_i L(z_i,\theta) + \eps L(z,\theta)
\end{align}
We can use calculus to derive a general formula $\theta^*$ relating
changes in the location of an optimum (or local extremum) $\theta^*$ to changes in a
second function parameter $t$. Suppose we have a function $f(\theta,
t)$ which we wish to minimize with respect to $\theta$:
\begin{align}
\theta^*(t)=\argmin_\theta f(\theta,t)
\end{align}
We are interested in knowing $\dby{}{t}\theta^*(t)$, in other words
how much the optimal $\theta$ will change when we vary the
``hyperparameter'' $t$. Under certain smoothness assumptions, we have,
from the stationary property of the optimum,
\begin{align}
\partby{}{\theta} f(\theta^*,t)=0.
\end{align}
Taking the derivative with respect to $t$ and applying the chain rule, we get
\begin{align}
\dby{}{t}(\ditto) = \partbyt{}{\theta} f(\theta^*,t) \dby{\theta^*}{t} + \partbyby{}{\theta}{t}f(\theta^*,t) = 0
\end{align}
Rearranging and multiplying by the ``inverse Hessian matrix'' gives
\begin{align}
\dby{\theta^*}{t}=-\(\partbyt{f}{\theta}\)^{-1} \partbyby{f}{\theta}{t} = -H^{-1}v
\end{align}
where we have introduced $H \equiv \partbyt{f}{\theta}$ and
$v\equiv\partbyby{f}{\theta}{t}$. This is the same as the new location
of the minimum if a quadratic function $\theta^\T H \theta$ were
perturbed by adding a linear term $\theta^\T v$ to it.\footnote{It
  should also be reminiscent of the update step in Newton's Method,
  $\theta_{t+1}=\theta_t - (\partbyt{f}{\theta}(\theta_t))^{-1}
  \partby{f}{\theta}(\theta_t)$, which also requires computing the
  inverse Hessian and multiplying it by a gradient vector.}

The standard name for this equation is the ``implicit function
theorem'' - although it might be more properly called the ``extremum
sensitivity formula'' since it calculates the sensitivity of extrema to
hyperparameters - and it is usually credited to Cauchy. It is
useful for approximating the effect of changing various aspects of a
regression problem, for example adding a new data point. Rather than
retraining the model for each new candidate point, we can represent
the addition as a small perturbation in the empirical risk function
and make a linear approximation using derivatives. Later on, we devote
section \ref{sec:comp-infl} to calculating $-H^{-1}v$, but first we go over some ways of
using these quantities in learning.

Define
\begin{align}
\theta(\eps,\zu) = \argmin_\theta \( f(\theta,z,\eps) \equiv R(s,\theta)+ \sum_i L(z_i,\theta) + \eps L(\zu,\theta) \) \label{eq:up-z-eps}
\end{align}
Then, loosely following\footnote{Our definitions are
  slightly different. Our $\eps$ is $\frac{1}{n}$ of theirs, so our
  $\I$ will be $n$ times larger; but the distinction is unimportant
  here.} Koh and Liang \cite{koh2017}, define
\begin{align}
\Iup(\zu) = \left. \dby{\theta^*(\eps,\zu)}{\eps} \right\vert_{\eps=0} = -H^{-1}\partby{}{\theta} L(\zu,\theta^*)
\end{align}
This approximates the change in the parameters caused by adding a new
data point $z$ to our data set. Similarly, we can approximate the
effect of adding $z$ on the loss at a given point $\zt$:
\begin{align}
  \Iul(z,\zt) &= \left. \dby{L(\zt,\theta^*)}{\eps} \right\vert_{\eps=0}  \label{eq:iul}
  = \partby{L}{\theta}(\zt,\theta^*)^\T \left. \dby{\theta^*}{\eps} \right\vert_{\eps=0} \\
  &= \partby{L}{\theta}(\zt,\theta^*)^\T \Iup(z) = -\partby{L}{\theta}(\zt)^\T H^{-1} \partby{L}{\theta}(z)
\end{align}
where we have omitted the $\theta^*$ parameter for
brevity.

Koh and Liang explore how to use influence functions to
perform tasks like identifying errors in a data set (``data set curation'').

Assuming a symmetrical $H$, from the final form of $\Iul$ we can see
that it is commutative, in other words invariant to swapping $z$ and
$\zt$.

\subsubsection{For cross-validation} \label{sec:for-cv}

Here is a more specific example of how we can apply these influence function
formulae to the analysis of a machine learning task. We can
approximate the generalization error, which is the objective for
optimizing the regularity $s$ in Leave-One-Out cross validation
(LOOCV) (equation \ref{eq:k-fold-reg-opt}), as proportional to
\newcommand{\Gcv}{G_{\textrm{CV}}}
\newcommand{\Gpert}{G_{\textrm{pert}}}

\begin{align}
  \Gcv(s) = \sum_{j=1}^n L(z_j,\theta^*_j(s)) \label{eq:gcv}
\end{align}
where
\begin{align}
  \theta_j^* = \argmin_\theta \sum_{i\neq j} L(z_i,\theta)
\end{align}
Now
\begin{align}
  L(z_j,\theta_j^*(s)) &\approx L(z_j,\theta^*)-\Iup(z_j)\cdot \partby{}{\theta} L(z_j) \\
  &= L(z_j,\theta^*) + \gl_j H^{-1} \gl_j
\end{align}
where we defined $\gl_i \equiv \partby{L}{\theta}(z_i,\theta)$. So
\begin{align} \Gcv &\approx \sum_{j=1}^n L(z_j) + \sum_{j=1}^n \gl_j H^{-1} \gl_j \end{align}
Since this approximation of the generalization error is a perturbation
on the (unregularized) empirical risk, we could name the approximation
$\Gpert$:
\begin{align} \Gcv(s) \approx \Gpert(s) = f(\theta^*(s))+ \sum_{j=1}^n(\gl_j H^{-1} \gl_j)\big|_{\theta=\theta^*(s)} \label{eq:gpert}
\end{align}
where $f$ is the (unnormalized) empirical risk
$f(z,\theta)=\sum_{i=1}^n L(z_i,\theta)$. Each term $\gl_j^\T H^{-1}
\gl_j$ approximates a correction to the loss on data point $j$,
$L(z_j,\theta^*)$, when $\theta^*$ is adjusted to account for the
effect of removing point $j$ from the data set. We call this
$\gl_j^\T H^{-1} \gl_j$ term the {\em self-influence}.
The approximation in equations
\ref{eq:gcv}-\ref{eq:gpert} could be useful for approximating the
LOOCV-optimal regularity, which makes the fullest use of the data,
without explicitly retraining the model $n$ times. It seems that it is
not possible to efficiently calculate $v^\T H^{-1}v$ for arbitrary
vectors $v$, without calculating the full inverse Hessian $H^{-1}$,
but if we could, then we could approximate the LOOCV
optimum regularity (equation \ref{eq:k-fold-reg-opt}) as
\begin{align}
  s^* \approx \argmin_s \Gpert(s) \label{eq:loocv-opt-reg}
\end{align}
Note that $\Gpert$ cannot be seen purely as a function of $\theta$
because the summed self-influence term will generally depend on $s$
through $H$. It is nevertheless tempting to view this term as a kind
of regularizer since the other term is the (unregularized) empirical
risk. Also, although $\Gpert$ cannot be used to directly optimize
$\theta$, it might be useful for optimizing other hyperparameters (see
the footnote on page \pageref{my-comp-sens}). In the case of RLS, it
can be shown that the self-influence summation reduces to the trace of
a product of several matrices.

The formula in \ref{eq:gpert} may appear in \cite{golub1979}. It does
not seem to appear in the classic monograph on influence functions
\cite{cook1982}. We do not use it here except in a ``future work''
sense.

\subsubsection{In active learning} \label{sec:in-active}

Influence functions can also theoretically be used for {\em active
  learning} by helping us identify new data points to request labels
for \cite{liu2021,xia2023}. Here, active learning refers to the {\em
  semi-supervised learning} task of querying a human or other source of
data to provide labels for specially chosen unlabeled data points.
Semi-supervised learning refers to the machine learning domain where
there are some number of complete (``labeled'') data points, together
with a potentially much larger number of data points with some
dimension missing, which are called ``unlabeled'' (for example points
where $x$ is known but not $y$ in equation \ref{eq:ols}).

We could (naively) imagine choosing new data-points for which to
request labels by calculating the influence of each potential query
data point on the expected loss of the model over some set of test
data $T$,
\begin{align}
Q(z_i)&\equiv \sum_{j\in T} \Iul(z_j,z_i) = \gl_T^{\T} H^{-1} \gl_i \\
Q(x_i)&\equiv \E\left[Q(z_i=(x_i,y_i))\where p_y(y_i \where x_i,\theta^*)\right]
\end{align}
where $\gl_i\equiv \partby{L}{\theta}(z_i,\theta)$, and $\gl_T\equiv
\sum_{i\in T} \gl_i$, and $Q(z)$ measures the ``quality'' of a new
labeled point z by estimating the expected change in total loss over
the test set if $z$ were to be included in training. The ``unlabeled
query heuristic'' $Q(x)$ averages $Q(z=(x,y))$ over possible labels
$y$, where $P_y(y|x,\theta)$ is some model-induced probability
distribution over the labels $y$ of a data point $(x,y)$.

However, generally (the linear regression case is treated in the footnote\footnote{If the loss $L$
  represents a negative log-likelihood, then
  the distribution $p_y$ is given by
\begin{align}
  p_y(y|x,\theta) &= {P(x,y|\theta)\over P(x|\theta)} = {P(x,y|\theta)\over \sum_y P(x,y|\theta)} = {\exp(-\alpha L(x,y,\theta)) \over \sum_y \exp(-\alpha L(x,y,\theta))}
\end{align}
where $\alpha$ is an arbitrary scaling constant.

For ordinary least squares, this reduces to
\begin{align}
  p_y(\cdot|x,\theta)&= N(\theta^\T x, {1 \over 2 \alpha}) \\
\implies \quad  p_y(y|x,\theta)&= {\sqrt{\alpha \over \pi}} \exp\(\alpha(y-\theta^\T x)^2\)
\end{align}
in other words a normal distribution with mean $\theta^\T x$ and
uniform variance $\sigma^2 = {1 \over 2 \alpha}$. Then
\begin{align}
\gl_i \equiv \partby{L}{\theta}(z_i,\theta) = \partby{}{\theta}(\theta^\T x_i-y_i)^2=2x_i(\theta^\T x_i - y_i)
\end{align}
and
\begin{align}
Q(x_i) &\equiv \E[Q(z_i)]=\E[\gl_T^\T H^{-1} \gl_i | y_i\sim N(\theta^\T x, \sigma^2)] \\
&= \gl_T^\T H^{-1} \E [(\gl_i = 2x_i(\theta^\T x_i - y_i))|y_i\sim \ldots] \\
&= \gl_T^\T H^{-1} 2x_i\(\theta^\T x_i - \(\E [y_i|y_i\sim \ldots] = \theta^\T x_i\) \) = 0
\end{align}
i.e.~$Q(x_i)=0$ due to the symmetry of the normal distribution around
its mean. This is clearly a problem for using this query selection
methodology in active learning.}), the
distribution over labels $y$ will be centered around the minimum of
the loss function $L(x,y|\theta)$ in such a way that
$\partby{L}{\theta}$ has zero expectation.

To avoid this zeroing effect, we can make the expectation non-linear
by averaging something like the square of the anticipated or projected
change in test
loss, for example with selectors like the ``squared total influence'' (STI):
\begin{align}
Q(z_i) = \(\sum_{j\in T} \Iul(z,z_i)\)^2
\end{align}
or the ``sum of squared influences'' (SSI):
\begin{align}
Q(z_i) = \sum_{j\in T} \Iul(z,z_i)^2
\end{align}

Neither heuristic seems easy to motivate philosophically, since we are
interested in simply reducing the loss, rather than changing it up or
down. However, the quadratic objective does lead to a simple
analytical formula for the expectation in the linear least-squares
setting:
\begin{align}
Q(x_i) &= \E[(\gl_T^\T H^{-1} \gl_i)^2|y_i\sim \ldots] \\
&= (2\gl_T^\T H^{-1} x_i)^2 \(\E[(\theta^\T x_i - y_i)^2|\ldots] = \sigma^2\) = 4\sigma^2 (\gl_T^\T H^{-1} x_i)^2
\end{align}
where $\sigma^2 = {1\over 2\alpha}$ is the variance of the response
variables in equations \ref{eq:alpha-lsr}-\ref{eq:alpha-prob-fact}.
If we calculate $\gl_T H^{-1}$ in advance and store its value, then
$Q(x_i)$ can be computed as a cheap dot product (which is then squared
and scaled).

In summary, the idea of trying to choose new unlabeled data points
based on their potential ability to reduce loss on some test set
doesn't immediately suggest a useful active learning heuristic,
because we can't know if the change in loss will actually be negative until we
see the label; and its expectation over the label will generally be
zero. However, query heuristics that are based on taking the
expectation of the square (or some positive function) of one or more
influence functions could be seen as measuring the potential
disruptiveness of a new data point in terms of its potential to change
the model's predictions, which intuitively seems like it could be
a useful heuristic in guiding an active learning algorithm.

The SSI ``sum of squared influences'' heuristic looks more useful at
first glance, since it captures the change in the model's response to
every data point, and is not, like the STI ``squared total influence'',
prone to ignoring changes in one loss that happen to be compensated by
opposite changes in another. However, it is more computationally
expensive because it requires a new influence function to be
calculated and stored for each data point in the test set $T$. The STI
heuristic by contrast only requires a single influence function $\gl_T
H^{-1}$ due to linearity: the total influence is
\begin{align}
\sum_{j\in T} \Iul(z,z_j) = \sum_j \gl_{z_j} H^{-1} \gl_z =
\(\sum_j \gl_{z_j}\) H^{-1} \gl_z = \gl_T H^{-1} \gl_z
\end{align}

If we extend the test set $T$ to encompass the entire data set, we
find that the stationarity of the optimal parameter vector gives
another interpretation of the total influence. Postulating a scaled
regularizer $R(s,\theta)=sR(\theta)$ in equation \ref{eq:gen-reg-reg},
we have:
\begin{align}
f(\theta,s)=\sum_j L(z_j,\theta)+sR(\theta) \label{eq:sr-reg-reg}
\end{align}
Differentiating by $\theta$ and setting the result to zero, we get
\begin{align}
  0 = \left. \dby{f}{\theta}\right|_{\theta^*=0}
  &= \sum_j \partby{L}{\theta}(z_j) + s \partby{R}{\theta}(\theta) \\
  &= \gl_T + s \rho  \\
\implies  \gl_T &= -s \rho
\end{align}
where we have defined $\rho = \partby{R}{\theta}$, which we call the
complexity gradient. The regularity $s$ can be compared with the
``upweighting'' hyperparameter $\eps$ from (\ref{eq:up-z-eps}). The
complexity gradient $\rho$ arises when computing the influence of the
regularity $s$ and is defined in the next section with respect to the
more general regularizer term $R(s,\theta)$, but the foregoing
observations allow us to reinterpret the squared total influence for
the whole data set as an influence function of the regularity term via
$s$: $\dby{\theta^*}{s} = - \rho H^{-1} = {1\over s} \gl_T H^{-1}$.
The value of this influence function at a data point $z$ is the ``loss
derivative'' $\dby{L(z,\theta^*(s))}{s}$ with respect to the
regularity, as $\gl_T H^{-1} \gl_z =
\dby{\theta^*}{s}\partby{L(z,\theta)}{\theta}=\dby{L}{s}$ (see next
section).

So we can see that
this leads to a query heuristic that prioritizes points at which the
model's responses vary most greatly when we adjust the regularity $s$
to make the model slightly over- or under-fit the data. Thinking in
terms of the influence of the regularizer seems to us to yield a more
intuitive viewpoint than thinking about sums over point-influences,
and it makes it more obvious that we are only calculating a single quantity.

Another line of reasoning that points us towards considering the
regularity of a model in active learning, is that for large models an
efficient query heuristic is only going to have time to calculate
something like an inner product with each prospective data point's
loss gradient. The loss gradient has $\theta$ in the denominator
(contravariant vector) so we are looking for a quantity with $\theta$
in the numerator (covariant vector). This is satisfied by an inner
product that corresponds to applying the chain rule $\dby{L}{t} =
\sum_j \partby{L}{\theta_j} \dby{\theta_j}{t}$ for some variable $t$.
A natural choice for $t$ is the regularity $s$.

\pagebreak

\section{Observations} \label{sec:obs}

\subsection{Regularity tangents} \label{sec:reg-tans}

We have now, we hope, laid enough of a foundation to introduce and
motivate our proposed influence-based selection criterion for
active learning. Rather than using the change in optimal parameter
values $\theta^*$ induced by the up-weighting of a given data point,
we can look at how $\theta^*$ changes when we adjust the regularity
hyperparameter $s$:
\begin{align}
\dby{\theta^*(s)}{s} = -\(\partbyt{f}{\theta}\)^{-1}\(\partbyby{f}{\theta}{s}\) = - H^{-1} \rho \label{eq:theta-change}
\end{align}
where the ``complexity gradient'' $\rho \equiv \partbyby{f}{\theta}{s} =
\partby{}{\theta}\(\partby{R}{s}(s,\theta)\)$ depends only on the regularization term of
the objective $f$ due to the partial derivative with respect to $s$. ($R$ is from
equation \ref{eq:gen-reg}.)
For $\L_2$ regularization, $\rho$ will be a vector proportional to
$\theta^*$ (namely, $\rho=2\theta^*$).

This quantity has the advantage of being global over the data set, so that it only has to be calculated once.
Computing $-H^{-1} \rho = \dby{\theta^*}{s}$ will allow us to quickly calculate
for any given data point $z$ the quantity
\begin{align}
\dby{L(z,\theta^*)}{s} = \(\partby{L(z,\theta^*)}{\theta}\)^\T\cdot \dby{\theta^*}{s}
\end{align}
which is a simple inner product with the loss gradient. A selection criterion for active
learning can then be formed by taking the expectation of the square of
this ``loss derivative''; following the rationale of the previous section \ref{sec:in-active}:
\begin{align}
Q(z)&=\(\dby{L(z,\theta^*)}{s}\)^2 \\
Q(x)&=\E\left[Q(z) \where p_y(y \where x,\theta^*)\right] \label{eq:rt-sel-q}
\end{align}

It seems appropriate to call derivatives with respect to the
regularity ``regularity tangents'' as for example $\dby{\theta^*}{s}$
is a tangent vector to the (one dimensional) curve $\theta^*(s)$.
Alternatively, another name for the quantity ``the derivative of $b$
with respect to $a$'' is ``the adjoint of $a$ with respect to $b$'',
so we may refer to derivatives with respect to the regularity
hyperparameter as ``regularity adjoints''. The term ``regularity
tangent'' suggests a vector rather than a scalar, so perhaps we could
adopt both terms and say that an element of the regularity tangent
vector is a regularity adjoint.

Finding a new data point which maximizes the expected change in
squared derivative of the loss at that point with respect to the
regularity hyperparameter can be justified intuitively in the
following way.
\begin{quote}
Hypothesis: {\em The data points which have the greatest potential
  to stabilize the model are those for which the model's
  estimates vary most greatly as a function of model regularity.}
\end{quote}
In other words these are the points whose model predictions are most
affected when we allow the model to overfit or underfit the existing
data. This heuristic can also be motivated by noticing that it selects
the data point whose inclusion would produce the greatest expected
squared change in model complexity. This is because we can show
mathematically that $\dby{L}{s}(z,\theta^*) =
\dby{}{\eps}\(\partby{}{s}R(s,\theta^*)\)$, with the $\eps$
up-weighting of a data point $z$ as in equation (\ref{eq:up-z-eps}),
and we argued in section \ref{sec:bayes-cv} that
$\partby{}{s}R(s,\theta)$, at least in the usual case, imposes a
complexity ranking upon models $\theta$. We proceed to prove this
equality.

For regularized regression problems, using the symmetry of the Hessian
$H$,
\begin{align}
\dby{L(z,\theta^*)}{s} &= \(\partby{L(z,\theta^*)}{\theta}\)^\T\cdot \dby{\theta^*}{s} = -\gl_z^\T H^{-1} \rho \\
&= -\rho^\T H^{-1} \gl_z = \rho^\T \cdot \( \Iup(z)\equiv \left. \dby{\theta^*(\eps,z)}{\eps}\right|_{\eps=0} \) \\
&= \(\partby{}{\theta}\partby{R}{s}(s,\theta^*)\)^\T \left.\dby{\theta^*}{\eps}\right|_{\eps=0}
= \left. \dby{}{\eps}\(\partby{R}{s}\)\right|_{\eps=0}
\end{align}
So our selection criterion can be seen as choosing the point whose
inclusion in the data set would be estimated to produce the maximum
expected squared change in the regularization term $\norm{\theta}^2$.
Imitating the pre-existing nomenclature, we can give a new name to
this influence function:
\begin{align}
\Iur \equiv \left. \dby{R_s(\theta^*)}{\eps}\right|_{\eps=0}
\end{align}
where $R_s(\theta)\equiv \partby{R}{s}(s,\theta)$. Calculating $\Iur$
using regularity adjoints is more efficient than calculating this
quantity naively using $\Iup(z_i)$ for $i=1\ldots n$, as in our method
$-H^{-1}\rho$ need only be computed once, and then we just want its
inner product with the loss gradient at $z_i$,
$\partby{L}{\theta}(z_i,\theta^*) \equiv \gl_i$, which yields the
vector-inverse-Hessian-vector product $-\gl_i^\T H^{-1} \rho$ for each
data point of interest.\footnote{The equivalence of
  $\dby{L(z,\theta^*)}{s}$ and $\dby{R_s(\theta^*)}{\eps}$ can be seen
  as a special case of a more general duality relationship. Given an
  optimization problem
\begin{align}
\theta^*(a,b) = \argmin_\theta F(\theta,a,b)
\end{align}
we have the following identity:
\begin{align}
\dby{}{b}\partby{F}{a}(\theta^*,a,b) = \dby{}{a}\partby{F}{b}(\theta^*,a,b)
\end{align}
because the first expression is
\begin{align}
\dby{}{b}\partby{F}{a} &= \partbyby{F}{a}{\theta}^\T \dby{\theta^*}{b} = -\partbyby{F}{a}{\theta}^\T (H^{-1}) \partbyby{F}{b}{\theta} \\
&= \dby{}{a}\partby{F}{b}
\end{align}
as the Hessian is theoretically a symmetrical matrix. A further
specialization is when $F(\theta,a,b)=a M(\theta) + b N(\theta)$,
where we then have $\dby{M(\theta^*)}{b} = \dby{N(\theta^*)}{a}$.
Remember that partial derivatives ($\partby{}{a}$) hold $\theta^*$
fixed, while total derivatives ($\dby{}{a}$) consider it a function of
$a$.}

Note that the equivalence $\dby{L}{s} \equiv \dby{R}{\eps}$
holds for any form of regularizer $R$, which may be an arbitrary
function of $\theta$. A common variation on $\L_2$ regularization is
to only include certain parameters in the regularization term
$s\norm{\theta}^2$ (equations \ref{eq:ols} and \ref{eq:gen-reg}), for
example to ensure model responses that are unbiased in certain ways.
If the regularizer is $sR(\theta)=s\norm{\theta_M}^2 = s\sum_{i\in M}
\theta_i^2$ for some restricted set of parameter indices $M$ then
\begin{align}
\rho \equiv \partbyby{f}{\theta}{s} = \partby{R}{\theta}(\theta) = 2\theta_M \propto \theta_M \label{eq:rho-restr}
\end{align}
where $(\theta_M)_k = \left\{\begin{array}{rl} \theta_k & \textrm{ if
} k\in M \\ 0 & \textrm{ otherwise} \\ \end{array}\right.$. In this
case $-\gl_z^\T H^{-1} \rho$ is equal to $\left.
\dby{}{\eps}\right|_{\eps=0} \norm{\theta_M}^2$, so it now measures
the influence of $z$, i.e.~the effect of up-weighting $z$, on the new
regularizer.

\subsection{Regularity tangent derived query heuristics} \label{sec:rt-derived}

This section explores additional query heuristics aside from the ``squared
loss derivative'' heuristic introduced in section \ref{sec:reg-tans},
which may also make use of the regularity tangent. The body of this
section has been removed from this draft for reasons that we cannot
disclose.


\subsection{Influence functions and regularity tangents in an example regression problem}
\newcommand{\plotwidth}{0.8\textwidth}
\label{sec:examples}

This section presents some plots illustrating the calculation of the
squared loss derivative (SLD) query heuristic for a simple polynomial
regression problem with degree 5 (i.e.~having six parameters, which are
the coefficients of the polynomial).

\begin{figure}[h] \centering
  \includegraphics[width=\plotwidth]{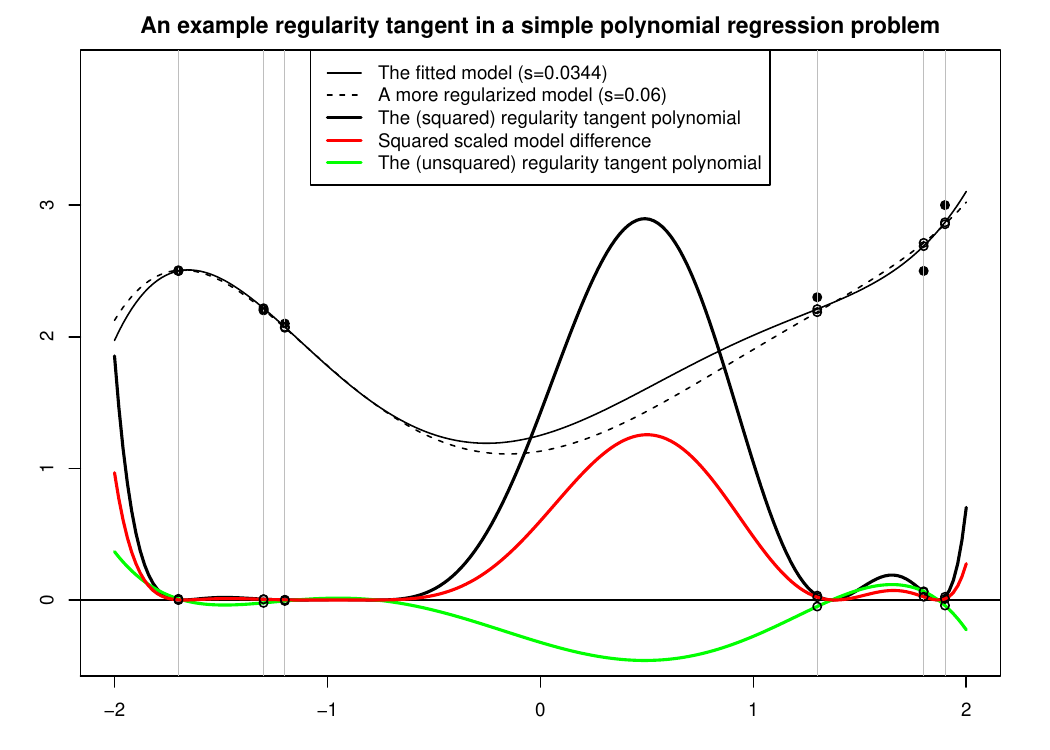}
  \caption{\label{fig:model-plot}}
\end{figure}

Figure \ref{fig:model-plot} shows the six data points (as filled
circles) which are used to train the model. These points have been
chosen so as to leave a large gap in the middle of the model domain. The
optimal regularity is determined using LOOCV ($s=0.0344$). The model
responses when trained at a slightly higher regularity ($s=0.06$) are
shown as a dashed line. The (squared) response from the regularity
tangent influence function is shown as a thick black line. Below it is
the approximation to the same quantity calculated as the difference
between the two model responses divided by the difference in
regularities, as a thick red line. The green line at the bottom shows
the response from the regularity tangent before it is squared, which
takes both positive and negative values.

\newcommand{\power}{\mathop{\mathrm{power}}}

The responses are calculated as\footnote{Here $F$ is the response
function notation from equation \ref{eq:nonlin-reg}, not to be
confused with the empirical risk $f$.}
\begin{align}
y(x) = F(x;\theta) = \power(x)^\T\theta
\end{align}
where $\power(x)=(1,x,\cdots, x^5)$ is a 6-element vector containing
powers of $x$ (or a 6-column matrix, if $x$ is a vector, with each row
corresponding to one data point). If $x$ is a vector of unlabeled data
points, and $y$ contains the labels, the optimal $\theta$ may be
calculated as
\eqntitle{The Regularized Normal Equations}
\begin{align}
\theta = (X^\T X + sI)^{-1} X^\T y
\end{align}
where $X = \power(x)$. We use the convention that $y$ and $\theta$ are
column vectors, and $X_i = \power(x_i)$ is a row vector with the
features of the $i$th data point.

This is the regularized version of
the well-known formula for the parameters of a linear regression
problem, called {\em the normal equations}, which incidentally takes
the familiar form of an inverse-Hessian-vector product, where $2X^\T y$
is the gradient at $\theta=0$ and $2(X^\T X + sI)$ is the (constant) Hessian. The
(regularized) normal equations are derived by setting the gradient of
the RLS objective function in equation \ref{eq:l2-reg} to zero:
\begin{align}
0 &= \dby{}{\theta}\( \sum_i (X_i\theta - y_i)^2 + s\theta^\T\theta\) \\
&= \sum_i 2 X_i (X_i \theta -y_i) + 2s\theta \\
0 &= \sum_i X_i \( \sum_j X_{ij} \theta_j - y_i \) + s \theta \\
(\forall k)\qquad 0 &= \sum_i X_{ik} \( \ditto \) + s \theta_k \\
&= \sum_{ij} \(X_{ik} X_{ij}\theta_j - X_{ik}y_i\) + s\theta_k \\
&= \sum_{ij} \(X_{ki}^\T X_{ij}\theta_j + s I_{kj}\theta_j\) - \sum_{ij} X_{ki}^\T y_i \\
X^\T y &= (X^T X + sI)\theta \\
\theta &= (X^T X + sI)^{-1} X^\T y \label{eq:reg-norm-eq}
\end{align}
It is equivalent to Newton's method in this model, which converges
after a single iteration because the objective function is quadratic.

In the normal equations, $X$ is a matrix with columns corresponding to
the ``features'' (in this case, powers of the $x$-coordinate) of each
data point, and rows corresponding to the data points. Thus $X^\T X$
is therefore a square matrix with the same dimension as the parameter
vector, $\dim \theta = p = 6$ in this example.

The regularity tangent is calculated as
\begin{align}
\dby{\theta^*}{s} &= -H^{-1}\rho = -(X^\T X + sI)^{-1}(2\theta^*)
\end{align}
One can verify that this is the same as calculating the $s$-adjoint of
\ref{eq:reg-norm-eq}.\footnote{Using the matrix-by-scalar differentiation rule
$\dby{U^{-1}}{s}=-U^{-1} \dby{U}{s} U^{-1}$, the derivative of equation \ref{eq:reg-norm-eq} becomes
\begin{align}
  \dby{\theta^*}{s} &= \dby{}{s}\((X^\T X + s I)^{-1} X^\T y\) \\
   &= \(\dby{}{s}(X^\T X + s I)^{-1}\) X^\T y \\
   &= -(X^\T X + s I)^{-1} \dby{(X^\T X + s I)}{s} (X^\T X + s I)^{-1} X^\T y \\
   &= -(X^\T X + s I)^{-1} \cdot I \cdot \theta^* \\
  &= -(X^\T X + s I)^{-1} \theta^* \\
  &= - H^{-1} \rho
\end{align}
}

\begin{figure}[H] \centering
  \includegraphics[width=\plotwidth]{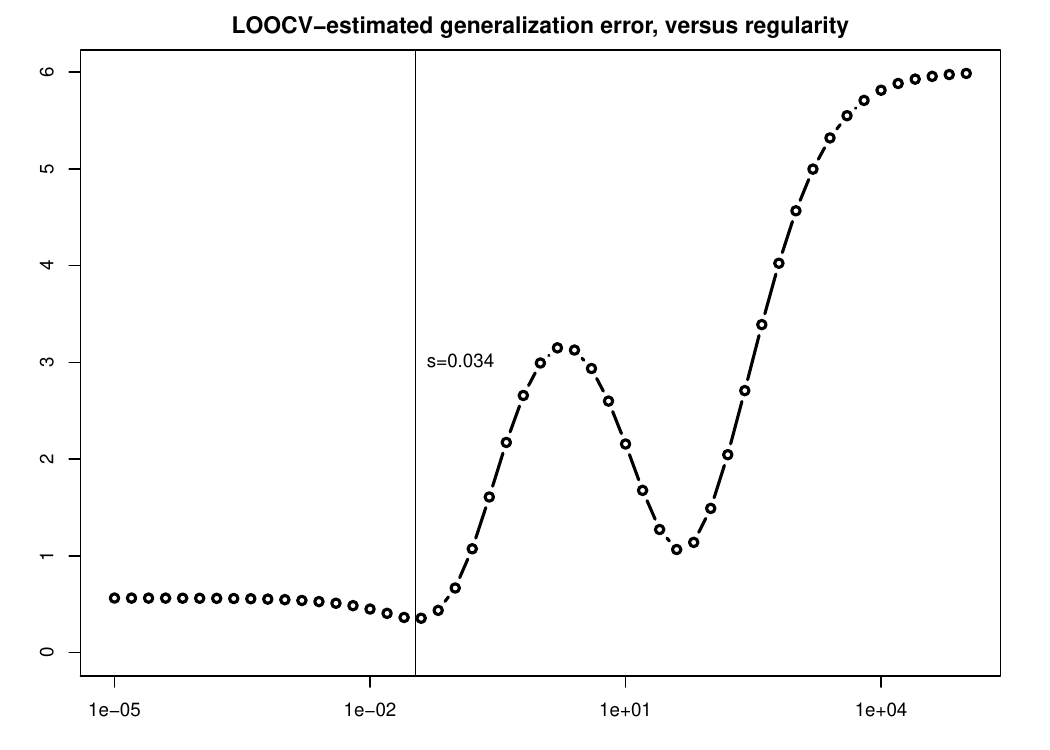}
  \caption{\label{fig:ge-plot}}
\end{figure}

Figure \ref{fig:ge-plot} illustrates how the optimal regularity was
calculated by minimizing the model's generalization error, which is
estimated using leave-one-out cross-validation (LOOCV). It is apparent
that local minima can exist, although in this example the fact is more
pronounced due to the small number of data points. It is also easy to
produce examples where the optimal regularity is outside of the range
of values we show in this plot, in other words it converges to zero or
infinity.

\begin{figure}[H] \centering
  \includegraphics[width=\plotwidth]{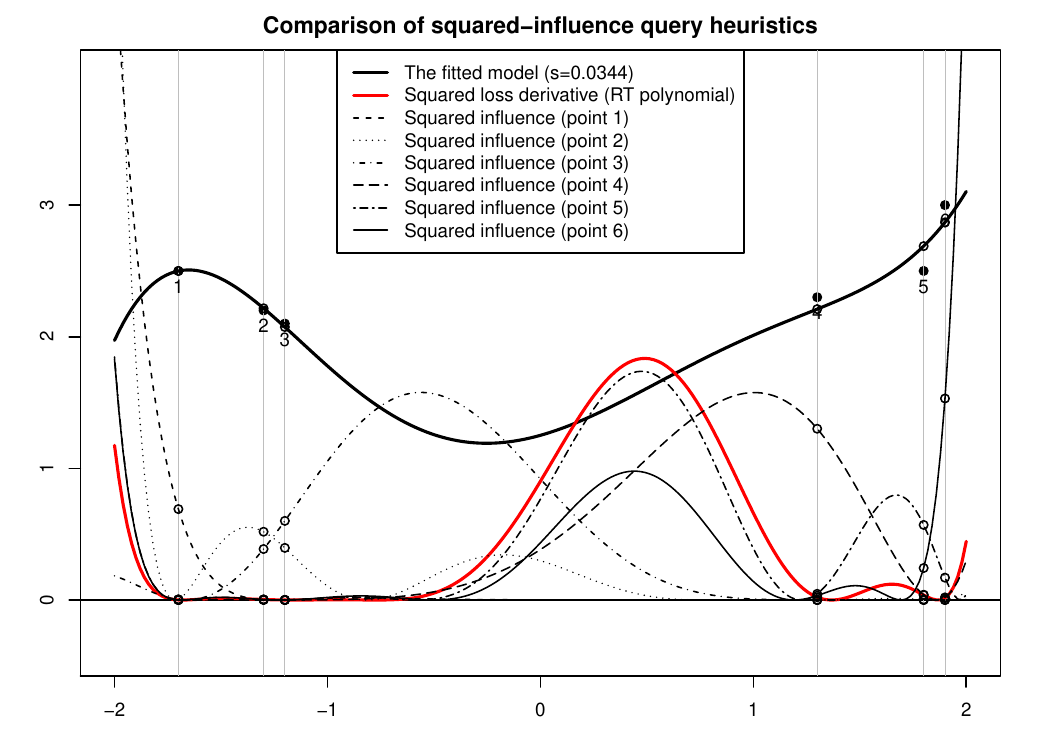}
  \caption{\label{fig:sis-plot}}
\end{figure}

Figure \ref{fig:sis-plot} compares the RT-based ($\Iur = \dby{R}{\eps}
= \dby{L}{s}$) SLD query heuristic with the squared-influence
heuristic derived from the traditional $\Iul$ ($=\dby{L}{\eps}$)
influence function.

\newcommand{\Iulu}{\I_{\overline{\textrm{up}}\textrm{,loss}}}
\newcommand{\Iuluu}{\I_{\overline{\overline{\textrm{up}}}\textrm{,loss}}}

The query heuristic curves are normalized to have the same RMS values,
so multiplication by a constant has no effect. In this model the loss
gradient contains a factor proportional to the vector of residuals $X\theta-y$
and we simply omit this factor when calculating the ``unlabeled''
version of the query heuristic, i.e.~when we don't know $y$. Rather
than comparing the influence between $z$ and $z'$
\begin{align}
\Iul(z,z') = -\gl_{z'}^\T H \gl_z
\end{align}
we are using an unlabeled version
\begin{align}
\Iulu(x,z') = -\gl_{z'}^\T H \bar{\gl}_x
\end{align}
where $\bar{\gl}_x = 2\power(x)$. This is within a factor of the
``doubly-unlabeled'' influence function
\begin{align}
\Iuluu(x,x') = -\bar{\gl}_{x'}^\T H \bar{\gl}_x
\end{align}
and thus equivalent when used in the SI (squared influence) query heuristic.

The factor becomes relevant when we are summing influence functions,
as the (singly) labeled SSI $Q(x) = \sum_{z} \Iulu(x,z)^2$ and STI $Q(x) =
\left(\sum_{z} \Iulu(x,z)\right)^2$ give lower weight to data points with small
residuals, while the (doubly) unlabeled SSI $Q(x) = \sum_{x'} \Iuluu(x,x')^2$
and STI $\left(\sum_{x'} \Iuluu(x,x')\right)^2$ do not depend on $y$ and so give
``equal'' weight to each point within the semantics of the model,
which is to say that the points are weighted independently of their
residuals. For this reason the multi-point influence based query
heuristics SSI (green) and STI (blue) come in labeled and unlabeled
versions.

The unlabeled versions appear more ``equitable'', covering a smaller
range of values and being slightly bounded away from zero. The labeled
versions are dominated by the influence of point 5, which has the
highest residual. The red curve showing the squared loss derivative
actually overlaps the blue curve ``square total influence (labeled)'',
but is shown scaled slightly higher so both curves can be seen on the
plot.

The reason for the curves to overlap was explained at the end of section
\ref{sec:in-active}. The fact
that, in a trained model, the gradient of the model objective with
respect to $\theta$ must be zero implies that a gradient of the
regularizer is equal to the negative of the sum of the loss gradients.
So, when the (labeled) influence functions for each data point are
added together, we obtain a vector proportional to the regularity tangent. Thus,
differentiating by the regularity to calculate the regularity tangent,
as we propose in this paper, can be seen as an efficient way of
calculating the ``total influence'' of a dataset.

\begin{figure}[H] \centering
  \includegraphics[width=\plotwidth]{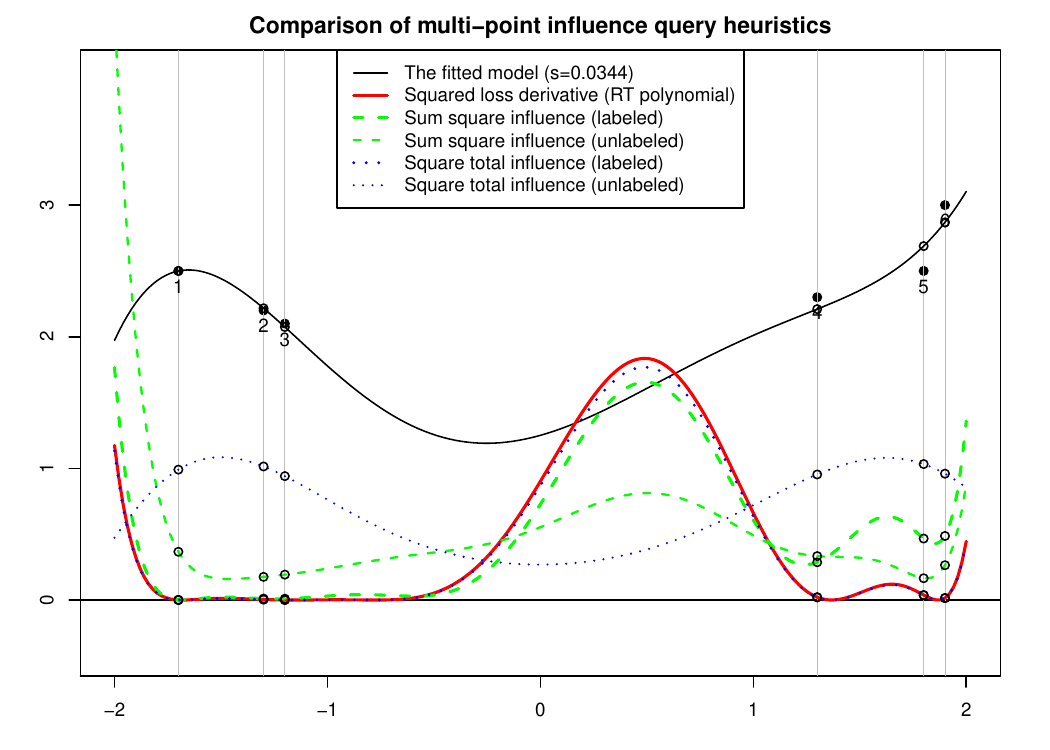}
  \caption{\label{fig:multi-sis-plot}}
\end{figure}

We hope that these plots have been helpful in visualizing the
quantities under discussion in this paper. They were generated using
R.

Obviously, future versions of this section should contain a comparison
of the various query heuristics under consideration, which at least
evaluates their suitability for active learning in the simple
polynomial regression model that we have been using here.

This is not expected to be difficult, but time constraints force us to
postpone such experiments for later. There is also the question of
testing the methodology in a more ``real world model'' setting, which
is where we hope it will be useful. This must also be regrettably
postponed, but the space of query heuristics that are theoretically
efficient enough for use in very large models seems sufficiently
constrained that one might hope the optimal heuristic to be among
those we have proposed in section \ref{sec:rt-derived}.

\subsection{Regularity in a multi-user setting}
\label{sec:reg-multi-user}


This section describes the hierarchical use of regularization in a
setting where individual users are attempting to train models that
exchange information by inheriting from a shared template model
containing all of the user interactions. This setting is important
because multiple users deserve multiple models, and therefore multiple
different parameter vectors and/or regularities; and yet it may be
desirable to combine information learned from each user so that data
and interactions are not wasted. Particularly when a user is new to
the system, we would like the first few interaction cycles to benefit
from knowledge that had been gained from other users, so that training
the model doesn't have to restart from scratch each time. We propose
two different approaches to the problem of building a multi-user
system out of an application incorporating a regularized regression
model.

\subsubsection{Multiple users via parameter inheritance}

A simple version of this idea might be to use the same regularization
coefficient for each user, but to use a regularizer that penalizes the
deviation of $\theta$ from some global parameter vector $\theta_0$,
rather than (as in the usual case) its deviation from zero:
\begin{align}
R(s,\theta) = s \norm{\theta-\theta_0}_1
\end{align}
We might suggest $\L_1$ regularization here because it enforces sparsity in
the deviation $\theta-\theta_0$, so that each user's $\theta$ differs
from the base model $\theta_0$ at only a finite number of indices.
Although $\L_2$ regularization is slightly simpler to implement
because the regularizer is smooth, $\L_1$ regularization could be
useful in the case of very large models, where storage considerations
prevent us from being able to make a full copy of the parameter vector
for each user. The shared parameter vector $\theta_0$ is optimized
using data from as many users as possible, perhaps using the same
regularity hyperparameter, and it will have its own regularity tangent
$\dottheta_0$. The regularity tangent $\dottheta$ for a single user
will inherit from the common $\dottheta_0$:
\begin{align}
\dottheta = \dby{\theta}{s} = \dby{\theta}{\theta_0} \left(\dby{\theta_0}{s}\right)_{(\theta)} + \left(\dby{\theta}{s}\right)_{(\theta_0)}
\end{align}
where $\left(\dby{\theta}{s}\right)_{(\theta_0)}$ means taking the
derivative of $\theta$ with respect to $s$ while considering
$\theta_0$ to be constant; we are considering $\theta$ to be an
(implict) function of both $\theta_0$ and $s$, while $\theta_0$ is an
(implicit) function of $s$ that is derived from the existing users
collectively in a separate training run.

Since each new user's $\dby{\theta}{\theta_0}$ starts out as the identity matrix,
this inheritance property from $\dottheta_0$ ensures that each user's model
has a non-zero regularity tangent even before any data points have
been seen, which is important because it means that the first query
shown to a user is based on the curiosity expressed by the common
model, and need not be random. It is not necessary to calculate the
full Jacobian matrix $\dby{\theta}{\theta_0}$, as our usual method
for calculating $\dottheta$ using (forward-mode) automatic
differentiation should yield it more directly (see next section).

The idea could be extended to use some kind of dimensionality
reduction on $\theta$ so that each user's deviation from $\theta_0$ is
described by a very small number of parameters, representing a
``personality type'' for example, which are learned over the course of
interacting with him or her. (An $\L_2$ regularizer would be
more suitable than $\L_1$ in this case.)

Returning to the $\norm{\theta-\theta_0}_1$ model, we note that for
very large models in practice the calculation of $\dottheta_0$ should proceed in a
very different way from the calculation of $\dottheta$. While
$\dottheta_0$ would be calculated stochastically using SGDF (which we
introduce in section \ref{sec:sgdf}) and batches of data points, the
interaction history of a single user would usually fit into a single
batch and so $\dottheta$ could be calculated using more efficient
second-order methods like NLCG (see section \ref{sec:second-order}),
which require a non-stochastic objective, presuming that these could
be adapted to work well with $\L_1$ regularization. It seems that this
or a similar hybrid optimization approach might be the best way to
give large language models a sense of curiosity about their data, in
addition to providing them with the ability to adapt to information
coming from an individual user, without necessarily increasing the
complexity of inference in the model. \label{sec:hybrid-approach}

\subsubsection{Multiple users through regularizer hierarchy}

We next explore the case where a multi-user model incorporates a
different regularizer for each user. This scenario is less interesting from the
perspective of modeling, but attempts to explore a natural
generalization to the regularization concept and to look at how it
interacts with the idea of regularity tangents and query heuristics.
Consider a model which contains certain user-specific parameters
$\theta_{M_i}$ (indexed by elements of the set $M_i$) for some set of
users indexed by $i$. We might find it interesting to use per-user
regularization coefficients $c_i$ whose distributions are controlled
by a single regularity hyperparameter $s$:
\begin{align}
  R(s,(c,\theta,\alpha)) = s \sum_i c_i^2 + \sum_i \( \alpha c_i \norm{\theta_{M_i}}^2 - {\abs{M_i} \over 2} \log \alpha c_i \) \label{eq:rsct}
\end{align}
In the Bayesian interpretation, the second term is saying that the
parameters for user $i$, namely $\theta_{M_i}$, are given a prior
consisting of independent normal distributions with variance ${1 \over
  2 \alpha c_i}$ and mean 0; see the discussion below equation
\ref{eq:alpha-lsr}. We multiplied the precision $c_i$ by $\alpha$ which
is to say half the Bayesian precision $2\alpha$ of the response
variables, following equation \ref{eq:aas}. We also gave
the ``pseudo-regularities'' $c_i$
normal distributions with mean 0 via the initial term.
The ${\abs{M_i} \over 2} \log \alpha c_i$ term is contributed from the
normalization constant of $P(\theta_{M_i}|c_i)$, which cannot be
ignored because it depends on $c_i$ which (unlike $s$) is a new model
parameter in $\theta$.
This hierarchical ``regularizer'' might be
used with a system that has data coming from different users and wants
to employ a slightly different prior when optimizing over each user's
parameter space, thus preferring simpler models for some users than
others. The prior distribution over the sub- or pseudo-regularities $c_i$ is
controlled by $s$, so that there is still only one regularity
hyperparameter ($s$) to be optimized outside of the normal (gradient
descent) training of $\theta$.

In equation \ref{eq:rsct} we should properly consider the
pseudo-regularizer
coefficients $c_i$ as elements of the vector $\theta$, with say
$\theta_{\hat{c}_i}\equiv c_i$ for each user $i$, where $\theta_{\hat{c}_i}$
denotes the index of $\theta$ being used to hold $c_i$. Then $\rho$ will only depend on $\theta_{\hat{c}} \equiv
\{c_i|i\}$:
\begin{align}
\rho \equiv \partbyby{f}{\theta}{s} \textrm{ ,\quad } \rho_j =  \left\{\begin{array}{lcl} 2c_i & : & j = \hat{c}_i \\ 0 & : & \textrm{otherwise} \\ \end{array}\right.
\end{align}
Even though $(\gl_z)_i$ may be zero except at the indices of certain
per-user parameters $i \in M_j$, where $z$ is a data point coming from
user $j$, and $\rho_i$ may be non-zero only at indices $\hat{c}_j$ not
intersecting any of the $M_j$, the regularity adjoint
$\dby{L(z,\theta^*)}{s}=\dby{\theta^*}{s}^\T\gl_z = -\rho^\T
H^{-1}\gl_z$ which combines these two vectors may still in general be
non-zero. That is due to couplings which will exist in the inverse
Hessian matrix $H^{-1}$ between the non-overlapping sets of parameter
indices. Thus although in this model our $\Iur(z)=-\rho^\T
H^{-1}\gl_z$ is computing $\dby{\norm{c}^2}{\eps}$, this quantity
itself depends indirectly on the per-user parameter change
$\dby{\norm{\theta_{M_i}}^2}{\eps}$ through $c_i$, where $i$ is the
user corresponding to data point $z$. Because $L(z,\theta)$ is only
connected to $s$ through $c_i$, we can write
$\theta^*(s)=\theta^*(c_i^*(s))$ and
\begin{align}
\dby{\norm{c}^2}{\eps} = \dby{L}{s} &= \sum_j \dby{L}{c_j} \dby{c_j}{s}
\end{align}
This is assuming there is no overlap between the per-user parameter
vector indices. Then,
\begin{align}
\dby{L}{s} =  \dby{L}{c_i} \dby{c_i}{s} &= \dby{\(2sc_i + \alpha\norm{\theta_{M_i}}^2-{\abs{M_i} \over 2 c_i}\)}{\eps} \dby{c_i}{s} \label{eq:multiuser-influence}
\end{align}
the long quantity being differentiated on the right-hand side being
$\partby{R}{c_i}$ because
\begin{align}
\dby{L}{c_i} = \dby{}{c_i}\partby{f}{\eps}=\dby{}{\eps}\partby{f}{c_i} = \dby{}{\eps}\partby{R}{c_i}
\end{align}

Thus for up-weighting a data point belonging to user $i$, which is
expected to only affect parameters in $\theta_{M_i}$, we have that the
regularity adjoint of the loss $\dby{L}{s}(z,\theta)$ is measuring the
effect on $\norm{c}^2$, equivalently $c_i^2$, of up-weighting point $z$. But this is also
approximately equal to the $\eps$ adjoint of a measure of the
complexity of the parameters $\theta_{M_i}$ specific to user $i$: $\alpha\norm{\theta_{M_i}}^2$.
(Here we have omitted the $2sc_i$ and $-{\abs{M_i} \over 2 c_i}$ terms that were in the numerator of (\ref{eq:multiuser-influence}), because these are constant with respect to $\eps$.)

This quantity is scaled by a user-specific constant $\dby{c_i}{s}$  in (\ref{eq:multiuser-influence}) which is part
of $\dby{\theta}{s}$, measuring something like our degree of interest
in user $i$. This scale factor does not affect the query heuristic,
and can be left in, as it is the same for every data point belonging
to a given user.


The above model can be represented by the probabilistic program, where
$N$ is the normal distribution:
\begin{align}
c_i &\gets N(0,{1\over 2s}) \label{eq:ci-draw} \\
\theta_{M_i} &\gets N(0, {1\over 2\alpha c_i}) \\
y_j &\gets N(\theta_{M_i}^\T x_i,{1\over 2\alpha})
\end{align}

In (\ref{eq:ci-draw}) above, the regularity $s$ could equivalently be
replaced with $\alpha s$, to make it look like (\ref{eq:draw-theta}),
but we could not then simply factor $\alpha$ out of the resulting
negative log likelihood objective because of the normalization
constants which depend on it nonlinearly.

The same idea could be extended to give each user $i$ a more complicated
prior over his parameters $\theta_{M_i}$, for example with a
system-wide mean $\theta_0$ and one or more principal dimensions of variation,
with $\theta_{M_i}^*(s)$ reacting to changes in $s$ through these
intermediate variables. As a start, we could consider for example
generating a linear regression model from the following probabilistic
program:
\begin{align}
\textrm{(Gamma parameters, $\theta$)} &\gets N(0,{1\over 2\alpha s}) \\
\kappa &\gets \Gamma(\ldots) \\
\theta^{(j)} &\gets N(\theta, {1\over 2 \alpha \kappa}) \\
c_j &\gets \Gamma(\ldots) \\
y_i &\gets N(\theta^{(o_i)\T} x_i, {1\over 2\alpha c_{o_i}})
\end{align}
This model gives each user $j$ a separate parameter vector
$\theta^{(j)}$ taken to be near some prototype $\theta^{(0)}$, with
precision controlled by $\kappa$, and furthermore assumes that each
user's data points are generated with a different precision,
controlled by $c_j$. The index $o_i$ refers to the user (``owner'')
for data point $z_i$. The Gamma prior over $\kappa$ and $c_j$ is the
conjugate prior for the precision of a normal distribution, and
therefore can be adjusted to capture the effect of simulated
observations.

In creating a regression problem for this model, the last ``sampling
operation'' $y_i \gets N(\ldots)$ gives rise to a sum of weighted
losses, with each loss term having weight $\alpha c_j$ for points
belonging to user $j$, which is in contrast to the previous setting
where each loss term had equal weight. The four earlier sampling operations
comprise the regularizer, which has multiple terms for each user. Each
loss also contains a term $-{1\over 2}\log \alpha c_i$, which
captures $\log Z$ in the last distribution. Here only $x$ and $y$
are observed variables; $\alpha$ is an arbitrary positive number and a
scale hyperparameter that does not enter into the model complexity.
There are a variety of ways to choose $\alpha$, which may for example
be optimized as a hyperparameter like $s$, or treated as a model variable with a Normal or
Gamma prior, or estimated directly from the data. The optimal value of
$\alpha$ is a measure of the variance of the true response variable
$y$ with respect to the model's predictions $\theta^\T x$. And $s$ is
the regularity hyperparameter. The rest of the variables are model
parameters to be optimized during (stochastic) gradient descent
(remembering to at least occasionally take into account the gradient
of the regularizer term, for example at the end of each pass through
the data). The hyperparameter $s$ should perhaps be optimized to
minimize a sum of the loss terms on a test data set, according to
whatever performance metric is desired of the model, including for
example reducing the weight of data points from users with low $c_i$,
parroting the loss term coefficients appearing in the new empirical
risk.

In summary, all of these regression problems seem amenable to the use
of regularity tangents, introduced in the previous section, because
regularity tangents have the ability to link various interdependent
model parameters to each other through the regularizer term. From
these brief algebraic investigations we would like to infer that our
method might be successfully applied to models with hierarchical
notions (meaning layered or structured) of model complexity, including
but not limited to those introduced in this section, rather than just to
the uniform regularizer model of section \ref{sec:reg-reg} (e.g.~(\ref{eq:l2-reg})).

\subsection{Computing influence functions} \label{sec:comp-infl}

We have been talking about influence functions and other quantities
which, according to the implicit function theorem, are constructed by
multiplying the negative inverse Hessian matrix with one or more
gradient vectors. We have so far set aside the question of calculating
the Hessian $H$ or inverse Hessian $H^{-1}$. Although we have been
motivated by efficiency concerns to propose approximations based on
differential calculus to estimate the effects of adding or removing a
data point from the corpus, so as to avoid the need for retraining the
model after each addition or subtraction, these approximations have
resulted in formulae involving the Hessian which can itself be
intractable to compute. If the number of parameters $p$ is very large
then the Hessian, which has $p^2$ entries, may even be impossible to
store in memory, let alone invert. This is the case for many
commonly-used models in machine learning. Some modern models are so
large that even the parameter updates of Stochastic Gradient Descent
would be prohibitively expensive but for the fact that the gradient
vectors $\partby{L}{\theta}$ are designed to be sparse through the use
of normally-zero activation functions like the ``ramp function''
$\max(x,0)$ in the network that defines the model. However, we can
show that it is possible to calculate influence functions efficiently
even in the case of such very large models, with the same time
complexity as training the model.

To explain this, some familiarity with automatic differentiation
is useful. Automatic differentiation is a family of methods based on
the chain rule of calculus for computing the derivatives of programs.
These methods generally fall into two classes: reverse-mode automatic
differentiation, also known as back-propagation; and forward-mode
automatic differentiation, which may be implemented using ``dual
numbers''.\footnote{A dual number is a pair of real numbers
  $(x,\dot{x})$ representing a quantity and its derivative with
  respect to a designated variable, say $t$, which may be written
  $x+\dot{x}\ud t$. They may be added and multiplied according to the
  laws of calculus, for example $(x+\dot{x}\ud t)+(y+\dot{y}\ud
  t)=(x+y)+(\dot{x}+\dot{y})\ud t$ and $(x+\dot{x}\ud
  t)\cdot(y+\dot{y}\ud t)=xy+(\dot{x}y+\dot{y}x)\ud t$ and propagated
  through differentiable functions, $f(x+\dot{x}\ud
  t)=f(x)+f'(x)\dot{x}\ud t$. Dual numbers are like complex numbers
  but where the imaginary element $i$ has $i^2=0$ rather than $i^2=-1$.}
Reverse-mode automatic differentiation computes the derivative of a
single output variable with respect to multiple input variables, while
forward-mode automatic differentiation computes the derivative of
multiple output variables with respect to a single input variable.
Both methods have a runtime proportional to the runtime of the
original program (implying that they all have the same {\em time complexities}),
although (unlike forward-mode, which doesn't change the memory complexity), reverse-mode automatic differentiation also
requires an additional amount of memory proportional to the runtime. A key observation is that
automatic differentiation can be applied to any program, even one
which is iterative. \cite{gilbert1992,domke2012,maclaurin2015,franceschi2017}

Reverse-mode automatic differentiation is typically used to
calculate the gradients $\partby{f}{\theta}$ in gradient descent and
$\partby{L(z)}{\theta}$ in stochastic gradient descent (SGD). We can
additionally apply forward-mode automatic differentiation to gradient descent
to obtain, together with the optimal parameter vector $\theta^*$, an
estimate of $\dby{\theta^*}{s}$ which is the regularity adjoint with
respect to $\theta^*$.

Writing $\dot{\theta}$ for $\dby{\theta}{s}$, we simply substitute the
``dual number'' $\theta+\dot{\theta}\ud s$ in the gradient descent
algorithm, equation \ref{eq:gd}, which becomes
\begin{align}
\theta_{t+1}+\dot{\theta}_{t+1}\ud s \gets
(\theta_t+\dot{\theta}_t\ud s)-\eta_t &\(\partby{}{\theta}f(\theta_t+\dot{\theta}_t\ud s)=\right. \label{eq:gd-dual-unexp}\\
&\left. \partby{}{\theta}f(\theta_t)+
\(\partbyby{}{\theta}{s} f(\theta_t)+
\partbyt{}{\theta} f(\theta_t) \cdot \dot{\theta}_t\)\ud s
\) \label{eq:gd-dual}
\end{align}
So the update of the dual part ($\ud s$ terms) becomes
\begin{align}
\dot{\theta}_{t+1} \gets \dot{\theta}_t - \eta_t (\rho+ H \dot{\theta}_t) \label{eq:dual-part-update}
\end{align}
where the complexity gradient
$\rho=\partbyby{f}{\theta}{s}=\partby{}{\theta}R_s(\theta_t)$ was
defined after equation \ref{eq:theta-change}, and the Hessian H is evaluated at $\theta_t$.
Note that this is a hybrid application of both forward-mode automatic
differentiation ($\dby{}{s}$) and reverse-mode automatic
differentiation ($\partby{}{\theta})$. The expanded second term
contains a so-called ``Hessian vector product'', which quantity can be
easily calculated through the use of dual numbers in the gradient
expression $\partby{}{\theta}f(\theta_t+\dot{\theta}_t\ud s)$. In fact
it is well known that Hessian-vector products may be calculated as
easily as the gradient of a function \cite{pearlmutter1993}. This may be
done using either forward-mode or reverse-mode automatic
differentiation, or numerical approximation, and methods for computing
a Hessian-vector product are available in popular
automatic-differentiation libraries. For example, using forward-mode
automatic differentiation ($\dby{}{\eps}$) after the reverse-mode
($\partby{}{\theta})$,
\begin{align}
\left.\dby{}{\eps}\right\vert_{\eps=0} \partby{}{\theta}f(\theta+\eps v)
= \partbyt{}{\theta}f(\theta)\cdot v = Hv
\end{align}
where $H$ is the Hessian $\partbyt{f}{\theta}$. Or using reverse-mode
automatic differentiation (twice):
\begin{align}
\partby{}{\theta}\(v^\T \partby{f}{\theta}\) = v^\T \partbyt{f}{\theta} = v^\T H \label{eq:rev-rev-vhp}
\end{align}
which seems to be the method used (at least recently) by the
popular PyTorch library. The
competing JAX library seems to be able to compute a Hessian-vector
product using both methods. Also see \cite{baydin2018}, \cite{baydin2022}.

Thus, although it is possible to produce a similarly instrumented
version of gradient descent using only reverse-mode automatic
differentiation via equation \ref{eq:rev-rev-vhp} (see
\cite{agarwal2017b}), the hybrid forward-reverse algorithm of equation
\ref{eq:gd-dual-unexp} seems simpler and faster if forward-mode is
available.

We have noticed in some simple experiments that
(\ref{eq:dual-part-update}) can be used as-is to calculate
$\dot{\theta}$, even alongside algorithms like Adam where $\eta$ depends on
the history of gradient vectors.

In the above algorithm, just as with the parameter vector
$\theta$, the parameter derivatives $\dot{\theta}$ may
be initialized to random numbers, or they may be primed with a
previously converged value. For well-conditioned problems, the final value of
$\dot{\theta}$ should be insensitive to the initial conditions. If it converges, from (\ref{eq:dual-part-update}) we can see that it should converge to a vector $\dottheta$ with $H
\dottheta = -\rho$, as expected from (\ref{eq:theta-change}).
It is of course possible to run the non-dual
version of gradient descent to convergence, and then use the resulting
$\theta^*$ as a starting point of the dual number algorithm, which
will leave $\theta$ stationary and only update the
derivatives $\dot{\theta}$ at each iteration.

Instead of computing the gradient of the whole objective function $f$
during each update of gradient descent, we could cycle through the
available labeled data points $z$ and use the loss gradient
$\partby{L}{\theta}(z)$ as a ``randomly sampled'' proxy for
$\partby{f}{\theta}$. This is the idea behind the popular ``stochastic
gradient descent'' (SGD) algorithm, defined in equation \ref{eq:sgd}.
As with gradient descent above, SGD can also be applied to dual
numbers, to yield an algorithm that estimates $\dby{\theta^*}{s}$
along with $\theta^*$. With this modification the SGD parameter
updates become (compare to equations \ref{eq:gd}, \ref{eq:gd-dual} and
\ref{eq:sgd}):
\eqntitle{SGDF loss updates, dual number form}
\begin{align}
\theta_{t+1}+\dot{\theta}_{t+1}\ud s \gets
(\theta_t+\dot{\theta}_t\ud s)-\eta_t &\(\partby{L}{\theta}(z_{u_t}, \theta_t+\dot{\theta}_t\ud s)=\right.\\
&\left. \partby{}{\theta}L(z_{u_t},\theta_t)+
\partbyt{}{\theta} L(z_{u_t},\theta_t) \cdot \dot{\theta}_t \ud s\) \label{eq:sgd-dual}
\end{align}
where the expanded form on the second line is missing the
$\partbyby{}{\theta}{s}$ term from \ref{eq:gd-dual} as $L$ does not
depend on $s$. We call this algorithm ``stochastic gradient descent
with forward-mode differentiation'' or SGDF. \label{sec:sgdf} The
update of the regularity tangent can also be written separately from that
of the parameters:
\eqntitle{SGDF loss updates, parallel form}
\begin{align}
\theta_{t+1} &\gets \theta_t-\eta_t \partby{}{\theta}L(z_{u_t},\theta_t) \\
\dot{\theta}_{t+1} &\gets \dot{\theta}_t - \eta_t \partbyt{}{\theta} L(z_{u_t},\theta_t) \cdot \dot{\theta}_t
\end{align}

We again assume that the regularization term in the definition of the
objective $f$ is taken into account intermittently, for example at the
end of each batch (equation \ref{eq:sgd-reg-batch}). For that update
we have
\eqntitle{SGDF regularizer update, dual number form}
\begin{align*}
\theta_{t+1}+\dot{\theta}_{t+1}\ud s \gets
(\theta_t+\dot{\theta}_t\ud s)-\eta_t \(
\partby{}{\theta}R(s,\theta_t)+
\(\partbyby{}{\theta}{s} R(s,\theta_t)+
\partbyt{}{\theta} R(s,\theta_t) \cdot \dot{\theta}_t \) \ud s\)
\end{align*}
or equivalently
\eqntitle{SGDF regularizer update, parallel form}
\begin{align}
\theta_{t+1}&\gets\theta_t-\eta_t \partby{}{\theta}R(s,\theta_t) \\
\dot{\theta}_{t+1} &\gets
\dot{\theta}_t-\eta_t \(\partbyby{}{\theta}{s} R(s,\theta_t)+
\partbyt{}{\theta} R(s,\theta_t) \cdot \dot{\theta}_t \)  \label{eq:sgd-dual-reg}
\end{align}
which in the case of $R(s,t)=s\norm{\theta}^2$ becomes
\eqntitle{SGDF regularizer update, $\L_2$, dual number form}
\begin{align}
\theta_{t+1}+\dot{\theta}_{t+1}\ud s \gets
(\theta_t+\dot{\theta}_t\ud s)-\eta_t \(
2s\theta_t+
\(2\theta_t+ 2s\dot{\theta}_t\) \ud s\) \label{eq:sgd-dual-reg-ltwo}
\end{align}
or equivalently
\eqntitle{SGDF regularizer update, $\L_2$, parallel form}
\begin{align}
  \theta_{t+1} \gets \theta_t - \eta_t(2s\theta_t) &= (1-2\eta_ts)\theta_t \\
  \dot{\theta}_{t+1} \gets \dot{\theta}_t-\eta_t \(2\theta_t+ 2s\dot{\theta}_t\) &=
   (1-2\eta_ts) \dot{\theta}_t - 2\eta_t \theta_t
\end{align}
The $\theta_{t+1}$ update is rightly called ``shrinkage'' because it
shrinks each component of $\theta$ by the factor $(1-2\eta s)$. The
$\dot{\theta}_{t+1}$ update also shrinks $\dot{\theta}$ by $(1-2\eta
s)$ but additionally subtracts a vector proportional to $\theta$; we
can see this is the same as assigning to $\dottheta$ a weighted
average of $\dottheta$ and $-\theta$.

\subsection{Equivalence of SGDF and LiSSA} \label{sec:equiv-sgdf}

Our ``SGDF'' algorithm computes $\dby{\theta^*(s)}{s}$, which as we
have seen can also be written as
$\dby{\theta^*(s)}{s}=-\(\partbyt{f}{\theta}\)^{-1}
\partbyby{f}{s}{\theta}=-H^{-1}v$ where $H=\partbyt{f}{\theta}$ is the
Hessian of $f$ and $v=\partbyby{f}{s}{\theta}$ is a gradient of the
regularizer term. It is straightforward to combine the two SGDF updates
defined above, in equations \ref{eq:sgd-dual} and \ref{eq:sgd-dual-reg}, so
that the regularizer is taken into account with every update. We show
that the resulting algorithm generalizes an existing algorithm called
LiSSA proposed by Agarwal in 2017 \cite{agarwal2017b}. The purpose of LiSSA is to compute
inverse-Hessian vector products $H^{-1}v$ from an objective function
$f$ that can be written as an average of loss functions calculated at
different data points, i.e.~$f(\theta)={1\over n}\sum_i
L(z_{u_i},\theta)$. The LiSSA algorithm requires that we have the
ability to compute Hessian-vector products for the Hessians of the
loss function evaluated at random data points, which as we have said
can be done easily using standard automatic differentiation libraries.
The LiSSA algorithm is based on the observation that
\begin{align}
v H^{-1} &= v { I \over I-(I-H) } = v\sum_{k=0}^\infty (I-H)^k \\
&= \(\((\ldots)(I-H)+v\)(I-H)+v\)(I-H)+v
\end{align}
where the expansion on the second line is an application of a
well-known trick for computing polynomials without
exponentiation.\footnote{For example $x^3+3x^2+3x+1=((x+3)x+3)x+1$.}

If $\h_t$ is our current approximation to $H^{-1}v$ then this gives the update
\eqntitle{LiSSA}
\begin{align}
\h_{t+1} \gets v+(I-H)\h_t
\end{align}
Since $H$ is an average of loss Hessians at each data point, we can
approximate it stochastically by substituting the loss Hessian at a
random data point. If we do this at every update, we get Agarwal's
``LiSSA-sample'' algorithm, which is usually called LiSSA:
\eqntitle{LiSSA-sample}
\begin{align}
\h_{t+1} \gets v+(I-\l_t)\h_t
\end{align}
where $\l_t$ is defined as the loss Hessian
$\partbyt{}{\theta}L(z_{u_t},\theta^*)$ where $u$ is the
(possibly repeating) sequence of random data point indices used for
each update. \cite{koh2017,agarwal2017b}

We now show how to introduce a step size $\eta$ into the LiSSA update
by scaling the objective function and the parameter vector. We then
show that the LiSSA updates with step size are equivalent to an
application of forward-mode automatic differentiation (dual numbers)
to SGD, as given in equations \ref{eq:sgd-dual} and
\ref{eq:sgd-dual-reg}, on an objective function with a specially
chosen regularizer term.

Scaling $f$ in the LiSSA algorithm would not be expected to change the
location of the optimum parameter vector $\theta^*$, but the Hessian
$H$ would be scaled by the same factor as $f$. As a first attempt, we apply LiSSA to an
objective $f$ which has been scaled by $\eta$. This yields the
following version of the algorithm, where $v$ has also been scaled by $\eta$ to keep the product $\h \approx H^{-1}v$ the same:
\begin{align}
- \h_{t+1} &= -\eta v-(I-\eta \l_t)\h_t \\
  &= -\h_t -\eta (v- \l_t\h_t) \label{eq:lissa-scale-one}
\end{align}
We can apparently get to the same equation by scaling $\theta$ by
$\sqrt{\eta}$, which yields $\theta' = \sqrt{\eta} \theta$, and
\begin{align}
H' = \partbyt{}{\theta'} f = \(\partby{\theta}{\theta'}\)^2 \partbyt{}{\theta}f = {1\over \eta} H
\end{align}
Then we must write $v'={v \over \eta}$ in order to keep the product
$H^{-1}v$ the same. The new ($\theta$-scaled) LiSSA update becomes:
\begin{align}
  -\h'_{t+1} &= -\eta v' - (I-\eta \l'_t)\h'_t \\
  &= -\h'_t -\eta (v'- \l'_t \h'_t) \label{eq:lissa-scale-two}
\end{align}
Both \ref{eq:lissa-scale-one} and \ref{eq:lissa-scale-two} are
equivalent to a stochastic gradient descent update with forward-mode AD
(SGDF, equation \ref{eq:sgd-dual}) on the dual parameter vector
$\dot{\theta}$ with step size $\eta$, where $-\h \approx -H^{-1}v =
-\(\partbyt{f}{\theta}\)^{-1}\partbyby{}{\theta}{s}f(\theta^*)$
corresponds to $\dot{\theta}$, and
$v=\partbyby{}{\theta}{s}f(\theta^*)$ is a gradient of the regularizer
term, what we have called the complexity gradient $\rho$; for $\L_2$
regularization it is $2\theta$. We can see that LiSSA must correspond
to a form of stochastic gradient descent where the regularizer term is
considered at each parameter update, rather than at the end of
batches, since $v$ is included in each update of $\h_t$. Since the
regularizer's contribution to $\dottheta$ is non-sparse, i.e.~it
touches every index, obviously for sparse loss-gradient models it is better to
perform this update only at suitably spaced intervals.

It is not clear whether the authors of the LiSSA algorithm understood
that a step size parameter could be introduced implicitly by scaling
the parameters or the objective function, or by comparison to SGDF, or
that the result would correspond to a form of SGD using forward-mode
automatic differentiation. To turn a problem framed for LiSSA, i.e.~to
compute $H^{-1}v$ for some $f$ and $v$, into a regularity adjoint
calculation problem, it seems sufficient to take
\begin{align}
f(\theta)={1\over n} \sum_i L(z_i,\theta) + R(s,\theta)
\end{align}
where $R(s,\theta)$ is chosen so that its gradient
$\partbyby{}{s}{\theta}R(s,\theta)$ is equal to $v$,
e.g.~$R(s,\theta)=s \theta^\T v$. It should be enough to set $s$ to
zero and only consider its adjoints, or $s$ can be chosen to reduce
overfitting, as we described in section \ref{sec:bayes-cv}. There are
comments in the published literature about LiSSA requiring careful
tuning or showing poor convergence when $H$ is badly conditioned (see
\cite{schioppa2021} or section 4.2 of \cite{koh2017}), and it seems
possible that one or more versions of the algorithm with a step-size
parameter as given here would have better behavior when run with
smaller step sizes, but we have not tested this hypothesis yet. There
is not an obvious way to choose the step size for LiSSA, but if SGDF
is optimizing the parameters $\theta$ at the same time, then the dual
components $\dot{\theta}$ will be updated with the same step size as
the parameters $\theta$, and so that design choice will already have
been made.

\subsection{SGDF and hyperparameter optimization} \label{sec:hyper-opt}

In section \ref{sec:for-cv} we proposed using influence functions to
approximate the LOOCV estimate of the regularized model's
generalization error on a set of data points. In this section we
return to the model selection theme and explore some possible uses for
regularity tangents outside of active learning.


As we have shown, our regularity loss derivative (or loss regularity
adjoint) $\dby{L(z,\theta^*(s))}{s}$ is an influence function in the
up-weighting sense, because it is theoretically equal to $\Iur$, or
the change in regularizer $R(s,\theta)$ when up-weighting a point
$z$'s loss by $\eps$; or in other words
$\dby{R_s(\theta^*(\eps))}{\eps}$. We now point out that our influence
function can also be used to optimize the regularity hyperparameter
$s$ during cross-validation. In fact this can be done simultaneously
with SGD(F). Simply divide the data set into two groups, a training
set and a test set. Pick a data point $z$ from the training set and
use $\partby{L}{\theta}(z,\theta)$ and the gradient of the regularizer
to update the parameters $\theta$ (with the current step-size in SGD,
keeping track of $\dby{\theta}{s}\equiv\dot{\theta}$). Then, pick a
random data point $z$ from the test set and use the (scalar-valued)
``gradient'' $\dby{L}{s}(z,\theta)$, calculated as
$\partby{L}{\theta}\cdot \dby{\theta}{s}$, to similarly update the
regularity $s$. In other words both $s$ and $\theta$ are updated using
stochastic gradient descent, with data points coming from two separate
pools, but the gradient for $s$ is calculated using the current
regularity tangent to avoid multiple retrainings.

We might hope this algorithm to produce a converged parameter vector
$\theta^*$ minimizing the loss on the training set over $\theta$,
simultaneously with a converged regularity $s^*$ such that
$\theta^*(s^*)$ maximizes with respect to $s$ the loss on the test set
when using parameters $\theta^*(s)$. This is the standard goal when
using cross-validation to optimize a hyperparameter such as $s$, and
it can be achieved using some of the techniques we have presented for
calculating $\dby{L}{s}$ with only a constant slow-down to the SGD
algorithm per update.

Furthermore, as a theoretical digression, we note that it should be
possible to modify this algorithm by moving points back and forth
between the test and training sets, as long as this rearrangement is
done sufficiently infrequently that $\theta^*$ can converge to its new
value when the training set changes, but not so infrequently that
$s^*$ shows excessive variation when the test set changes. Presumably,
meeting both criteria requires using widely separate step sizes to
update $s$ and $\theta$. It might also make sense to do these updates
in batches, for example following each training batch with a test
batch. One might hope that the resulting algorithm would be something
like $k$-fold cross-validation, averaging over multiple test/training
splits of the data while optimizing $s$.

We have not yet tested this idea but it is interesting to think about
a simple gradient update algorithm like SGD that is capable of
performing the same regularity hyperparameter optimizations as a
methodology based on cross-validation, but on a continuous basis
during a uniform training phase, keeping only a single copy of the
parameter vector and its regularity tangent in storage, along with all
the data points. It is also likely that this modified algorithm
would be inefficient due to the need to update $s$ sufficiently
slowly.

In either case it seems interesting to ask if the existence of a
continuous algorithm for jointly optimizing $\theta$ and $s$ could
lead to any insights about the linear algebra and differential
geometry behind what is happening in regularized cross-validated
regression, whether in the original or the modified (with migration of
points between training and test sets) algorithm, and what this could
teach us about model selection. It is tempting to recommend the use
of influence functions in the modified algorithm for making
compensatory adjustments to $\theta$ every time the training set
changes, but each such estimate requires a separate run of LiSSA/SGDF,
as the influence of a data point $z$ on $\theta$, which we called
$\Iup$, cannot be straightforwardly computed from the regularity
tangent.

Note that the optimal value of $s$ may change slightly every time
points are added or removed from the training data, resulting in a new
$\theta^*$ and a new value for the complexity $R_s(\theta^*)$. Active
learning and online learning algorithms could theoretically benefit
from a training methodology like this that optimizes $\theta$ and $s$
together, if having a precise realtime estimate of $s$ turns out to be
important to the application.

Many implementations of SGD do not use a regularizer but rely on early
stopping to prevent overfitting (as we mentioned in section
\ref{sec:gd}), with the decision to stop training usually made by
periodically evaluating the model on some test data set and looking
for the moment when the loss starts to increase. In such settings it
should be possible to set $s$ to some small value and use a dummy
regularizer like $s\norm{\theta}^2$ just for the purpose of running
the SGDF algorithm for e.g.~active learning selectors, as we
hypothesize that the calculation of $\dot{\theta}$ will still be valid
and the loss-derivative model-complexity influence-function identity
$\dby{L(z,\theta^*)}{s} = \dby{R_s(\theta^*(\eps))}{\eps}$ will still
approximately hold even if $s$ is non-optimal, because the identity
only depends on the identification of an optimum for $\theta$ and not
$s$.

Finally, we would like to highlight the generality of our algorithm and
the techniques presented here by making an appeal to the universality
of the regularization concept in continuous models, which is based on
the simple and fundamental philosophical argument that all models may be
ranked by complexity, so that a single scalar hyperparameter suffices
to compensate for overfitting.

\subsection{Second-order optimization methods} \label{sec:second-order}

We have been concentrating on gradient descent and its variations
because they are currently used to train the largest machine learning
models. However, where it is available, optimization methods that use second-order derivative
information can be much more efficient than first-order methods like
gradient descent. These methods generally seem to prefer a
deterministic objective, so are mostly applicable to embodiments where the
training data is small enough to fit into one ``batch'', so that one
may forego stochastic optimization when training the model. Such
situations may in fact be common in interactive settings, as the
entire history of a single user's interaction with an application, or
at least the salient parts of it, is likely to be small enough to fit
into one training batch on commodity hardware. Where the Hessian
is small enough to be inverted, Newton's Method may be used for
training the model.
In this complexity domain, since we can invert
$H$, we may of course compute the trained regularity tangent directly
as $\dottheta^*=-H^{-1}\rho$ (equation \ref{eq:theta-change}).

Even when the full Hessian may not be computed directly,
Hessian-vector products are generally available, and are
asymptotically no more expensive than evaluating the model objective.
Thus, provided that all the data fits in a batch, the Nonlinear
Conjugate Gradient (NLCG) algorithm \cite{shewchuk1994}, which uses
Hessian-vector products to determine step size and direction, is
always applicable. It is supposed to take $\dim \theta = p$ steps in
the ideal case of a quadratic objective.

Theoretically, the NLCG algorithm itself will produce a correct
regularity tangent as a byproduct of the optimization if we apply it
to dual numbers as we did in equation \ref{eq:gd-dual-unexp} for
gradient descent. However, in this domain of medium complexity it is
also possible to compute the regularity tangent (or any other
influence function) using the Conjugate Gradients method (CG), the
linear method upon which NLCG is based, which can be seen as computing
$H^{-1} \rho$ using Hessian-vector products. This RT computation can
be done after running the model parameter optimization to convergence
using ordinary (not dual number) arithmetic.

The second option may be more convenient, in other words calculating
$\dottheta$ only as a special step at the end of training, and it
avoids layering three different kinds of automatic differentiation.
These three algorithmic transforms are needed to automatically
propagate dual numbers through NLCG (already a second-order algorithm)
as the influence of the dual variable on the step size becomes
important here. In preliminary experiments it was not possible to
ignore that dependency and still get accurate regularity tangents, as
it had been with Gradient Descent and Adam (section \ref{sec:gd}).
Furthermore, we would expect CG on $\dottheta$ to have faster
convergence if we run it separately after the convergence of $\theta$,
since the algorithm won't be tracking a moving target if
$\theta=\theta^*$ is already fixed. This is what we would
speculatively recommend for medium-sized, i.e.~single-batch but
intractable Hessian, models where a deterministic method like NLCG can
be used but Newton's Method is unavailable due to the intractability
of computing or storing the Hessian in the larger parameter
space.\footnote{\label{my-comp-sens} An alternative to NLCG in
medium-sized models is to use dimensionality reduction through random
embeddings to approximate the model with a smaller model, having
tractable Hessian. Newton's method can then be used on the smaller
model. A possible algorithm for reducing the model dimension starts
with a random linear embedding of some tractable parameter space (say
with 300 variables) into $\R^p\ni\theta$. The linear embedding may be
offset by our current parameter estimate $\theta_t$, so that
$\theta_t$ is recovered when all the small-model variables are zero.
Do the following in a loop ($\ast$): The model is considered as now
being parameterized by this smaller space $\R^{300}$ and its Hessian
is calculated, a $300 \times 300$ matrix. Then, SVD or some other
method is used to reduce this to a smaller, say $150 \times 150$
matrix, that reproduces the $300\times 300$ Hessian as closely as
possible when projected back and forth into $\R^{300}$. After this
``compression'' of the model, we have an embedding of $\R^{150}$ into
$\R^{300}$ into $\R^p$, which we simplify by composition into an
embedding of $\R^{150}$ into $\R^p$. This gives us a 150-variable
model that approximates our original model via linear embeddings. Now
we add another 150 variables to this model, which are chosen as before
to have random linear embeddings into $\R^p$; their values can be
initialized to zero. Now we have a 300-variable model, as we did at
the start of the loop, although half of the variables were inherited
from the previous iteration. Keep returning to ($\ast$) until some
kind of convergence is achieved. The idea behind the ``compressions''
is that they allow us to keep the most important information we have
about the structure of the true Hessian, while at the same time making
room for additional random variable embeddings that may have a chance
to reveal new structure in the original model. We are not sure if this
idea has been described elsewhere, although it was partly inspired by
a half-remembered conversation about ``compressed sensing''. The
iterations of the dimensionality reduction process can be interleaved
with for example Newton updates in the reduced space, which leads to a
nice parameter-free second-order approximate optimization algorithm
for medium-sized models. Such a setup would also make it possible to
directly optimize a perturbed version of the generalization error
$\Gpert$, as the RHS of equation \ref{eq:gpert} becomes tractable in
the embedding model, and that objective could then be used to optimize
$s$ or perhaps the embedding hyperparameters. We have not tested any
of these ideas yet.}

Provided that second-order methods can be used effectively in
$\L_1$-regularized models, in spite of their discontinuities in the
gradient function\footnote{The use of second-order methods on
non-smooth optimization problems or ones with $\L_1$ terms seems to be
the subject of several papers, but here we are only interested in the
fact that it is possible.}, we again note the possibility (brought up
in section \ref{sec:hybrid-approach}) for a hybrid learning system
that trains a large model using (a) a large commodity corpus and
(multi-batch) stochastic gradient descent with sparse loss gradients
to obtain a global parameter vector $\theta_0$ which is then (b)
refined using sparse updates via a (for example)
$\norm{\theta-\theta_0}_1$ regularizer and a suitable second-order
method in a deterministic single-batch mode since the (relatively
short) interaction history with a given user fits into a single batch
of data points, making it feasible to have a deterministic objective. Such a
``dual-mode'' system seems like the only tractable way for a system
embodying a large language model (or other large model) to change its
beliefs and express curiosity in response to interactions with an
individual user, without adding to the time complexity of its
responses. The records of such interactions could then be collected
from multiple users and used to further refine the base model
$\theta_0$.

It seems that a curated data acquisition methodology based on
efficient active learning might lead to more responsive, personalized,
and efficient training of large AI models and more compact and
efficient knowledge representations. Furthermore it would be possible
to substitute an existing AI model for a human user, leveraging the
existing model to create a more compact version of itself, which could
then be further refined by human input, and so on. Given the reasoning
power of existing AI language models, it is somewhat frightening to
speculate about the possible experience of interacting with an
intelligent computer system that is {\em actually} able to be
interested in a user's beliefs and to express curiosity about them and
perhaps even uncover contradictions. Provided that it can be made to
work, there is furthermore the possibility for such a system to be
used in interrogations by authorities of all kinds, which could lead
to various abuses. The authors are however hopeful that humanity would
ultimately benefit from the systems embodied in these ideas because of
their potential for use in human education and learning.

\section{Conclusion}

Our goal has been to explore ways in which a standard machine learning
regression model can be made to express a notion of curiosity. We have
done this through the framework of Active Learning, in which a model
is made to select unlabeled data points whose (labeled) addition to
its training data would be somehow beneficial to the accuracy of the model.

In exploring this problem, we have introduced a concept and a quantity
called a ``regularity tangent'' (section \ref{sec:reg-tans}), which
can be used for data point selection in active learning (ibid.), as
well as for optimization of the regularity hyperparameter in models
with explicit regularization (section \ref{sec:hyper-opt}). We showed
that the inner product of the regularity tangent with the loss
gradient at a data point is equivalent to the ``influence'' of that
data point on the model complexity as defined by the model's
regularizer (section \ref{sec:reg-tans}). Then we outlined some
potentially useful active learning query heuristics that incorporate
the regularity tangent (section \ref{sec:rt-derived}). We illustrated
the regularity tangent and related concepts with examples from a
simple linear regression problem (section \ref{sec:examples}). We
discussed multi-user models where the regularizer penalizes deviation
from a shared ``global'' parameter vector, as well as the application
of regularity tangents to multi-user models where there is a hierarchy
of regularizers (section \ref{sec:reg-multi-user}). We showed how to
calculate the regularity tangent in large models using a new algorithm
called ``Stochastic gradient descent with forward-mode automatic
differentiation'', or SGDF (section \ref{sec:sgdf}), which we showed
to be equivalent to a generalization of an existing algorithm called
LiSSA (section \ref{sec:equiv-sgdf}). We also outlined how to compute
the regularity tangent in small and medium-sized models where other
optimization methods than gradient descent are feasible, which use
second order derivative information, such as Newton's Method and NLCG
(section \ref{sec:second-order}).

\section{Acknowledgments}

The author would like to thank Clayton Otey, David Duvenaud, Zoubin
Ghahramani, and David Liu for assistance and helpful feedback during
various stages of this paper; but especially to Zoubin, who also
supervised my master's thesis and thereby nurtured an early interest
in Active Learning and Bayesian Machine Learning.

\bibliographystyle{unsrt}
\bibliography{genbib}

\end{document}